\documentclass[10pt,twocolumn,letterpaper]{article}

\usepackage{authblk}

\usepackage[pagenumbers]{cvpr}
\usepackage{multirow}
\usepackage{array}
\usepackage{bbm}
\usepackage{graphicx}
\usepackage{amsmath}
\usepackage{amssymb}
\usepackage{booktabs}
\usepackage{bbding}
\usepackage{color}
\usepackage{appendix}
\usepackage[dvipsnames]{xcolor}

\definecolor{cvprblue}{rgb}{0.21,0.49,0.74}
\usepackage[pagebackref,breaklinks,colorlinks,citecolor=cvprblue]{hyperref}
\usepackage{xcolor}
\definecolor{DarkGreen}{rgb}{0.43, 0.68, 0.28}

\newcommand{\network}{f}
\newcommand{\netweight}{\theta}
\newcommand{\imagei}{\mathbf{x}_i}
\newcommand{\targeti}{\mathbf{y}_i}

\newcommand{\imagej}{\mathbf{x}_j}
\newcommand{\targetj}{\mathbf{y}_j}

\newcommand{\imagemixup}{\mathbf{\tilde{x}}_i}
\newcommand{\targetmixup}{\mathbf{\tilde{y}}_i}
\newcommand{\losscrl}{\mathcal{L}_{\text{crl}}}
\newcommand{\weightcrl}{\lambda_{\text{crl}}}
\newcommand{\lossmixup}{\mathcal{L}_{\text{mix}}}
\newcommand{\weightmixup}{\lambda_\text{{mix}}}
\newcommand{\losscls}{\mathcal{L}_{\text{ce}}}
\newcommand{\losstot}{\mathcal{L}_{\text{total}}}
\newcommand{\mixcoef}{m}
\newcommand{\betadist}{\beta}

\newcommand{\scorei}{\mathbf{s}_i}
\newcommand{\scorej}{\mathbf{s}_j}
\newcommand{\sgn}{\text{sign}}

\def\paperID{11637} %
\def\confName{CVPR}
\def\confYear{2024}

\title{SURE: SUrvey REcipes for building reliable and robust deep networks}

\author[1,2]{Yuting Li}
\author[3]{Yingyi Chen}
\author[4,5]{Xuanlong Yu}
\author[6]{Dexiong Chen\textsuperscript{$\dag$}}
\author[1]{Xi Shen\textsuperscript{$\dag$}}

\affil[1]{Intellindust, China}
\affil[2]{China Three Gorges University, China}
\affil[3]{ESAT-STADIUS, KU Leuven, Belgium}
\affil[4]{SATIE, Paris-Saclay University, France}
\affil[5]{U2IS, ENSTA Paris, Institut Polytechnique de Paris, France}
\affil[6]{Max Planck Institute of Biochemistry, Germany}
\makeatletter
\renewcommand\AB@affilsepx{, \protect\Affilfont}
\makeatother

\begin{document}
\maketitle
\renewcommand{\thefootnote}{\fnsymbol{footnote}}
\footnotetext[2]{Corresponding Author.}
\begin{abstract}

In this paper, we revisit techniques for uncertainty estimation within deep neural networks and consolidate a suite of techniques to enhance their reliability. Our investigation reveals that an integrated application of diverse techniques--spanning model regularization, classifier and optimization--substantially improves the accuracy of uncertainty predictions in image classification tasks. 
The synergistic effect of these techniques culminates in our novel SURE approach. We rigorously evaluate SURE against the benchmark of failure prediction, a critical testbed for uncertainty estimation efficacy. Our results showcase a consistently better performance than models that individually deploy each technique, across various datasets and model architectures. When applied to real-world challenges, such as data corruption, label noise, and long-tailed class distribution, SURE exhibits remarkable robustness, delivering results that are superior or on par with current state-of-the-art specialized methods.
Particularly on Animal-10N and Food-101N for learning with noisy labels, SURE achieves state-of-the-art performance without any task-specific adjustments.
This work not only sets a new benchmark for robust uncertainty estimation but also paves the way for its application in diverse, real-world scenarios where reliability is paramount. Our code is available at \url{https://yutingli0606.github.io/SURE/}.

\end{abstract}
\section{Introduction}
\begin{figure}
\centering
\includegraphics[width=0.95\linewidth]{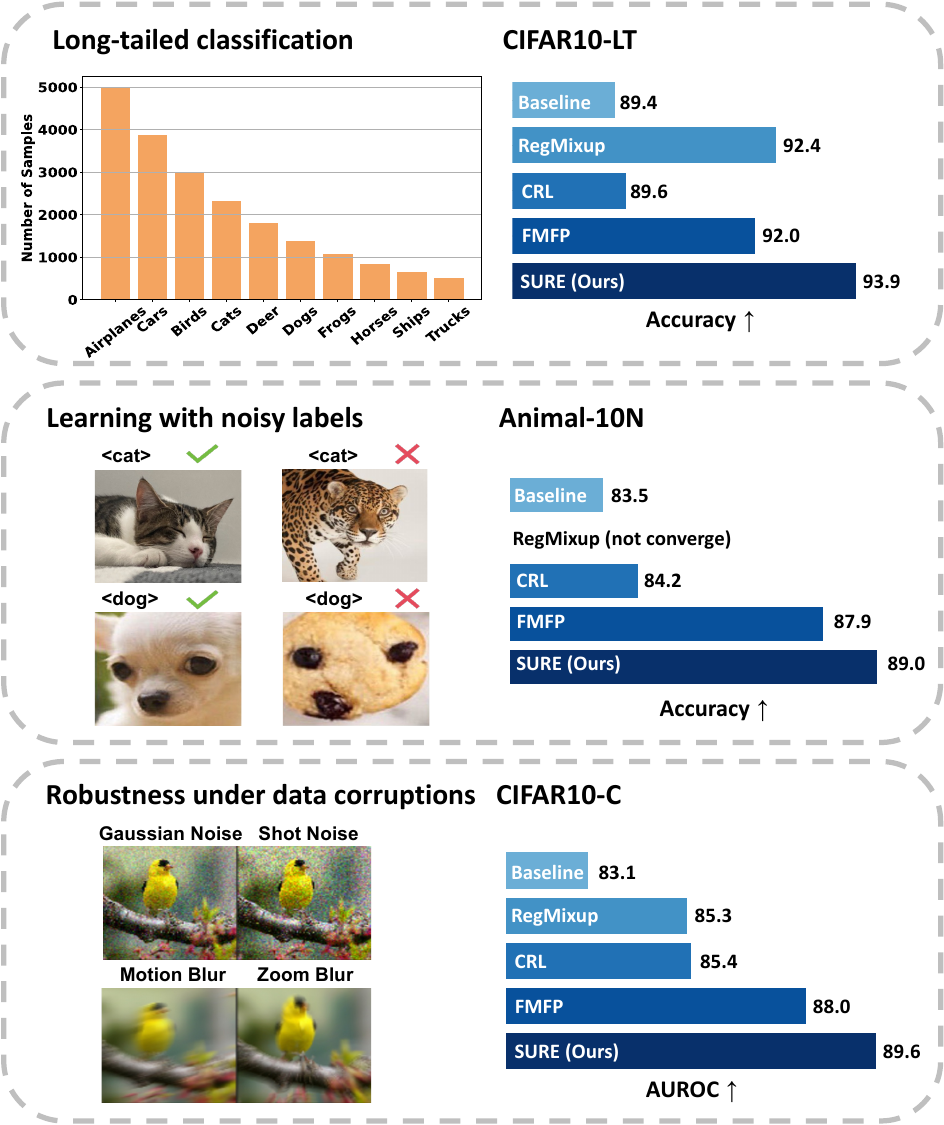}
\caption{\textbf{SURE consistently performs better than previous approaches to uncertainty estimation under various scenarios.} Note that we did not manage to scale RegMixup~\cite{pinto2022using} to the learning with noisy label task. Baseline refers to the MSP~\cite{hendrycks2016baseline} method.} 
\label{figure::teaser}
\end{figure}

Deep neural networks (DNNs) have established themselves as powerful and adaptable tools for prediction tasks on structured data. However, accurately assessing the reliability of their predictions continues to be a substantial challenge. In safety-critical areas such as medical diagnostics~\cite{leibig2017leveraging, ayhan2020expert, nair2020exploring}, robotics~\cite{loquercio2020general, he2015adaptive}, autonomous driving~\cite{choi2019gaussian, feng2018towards, loquercio2020general}, and earth observation systems~\cite{miralles2023critical,  giuffrida2021varphi}, decisions based on overconfident predictions can result in severe consequences. 
Consequently, ensuring the robust dependability of artificial intelligence systems grounded in DNNs is of utmost importance.

Addressing the issue of overconfidence in deep learning has been a focal point of significant research efforts, such as~\cite{guo2017calibration, liang2017enhancing, hendrycks2018deep, steinhardt2016unsupervised, liu2020energy, Mukhoti_2023_CVPR}. However, a key limitation of these methods is their restricted testing scenarios, typically confined to benchmark datasets for a single, predefined task like failure prediction or out-of-distribution (OOD) detection. The effectiveness of these methods in more complex, real-world situations involving issues like data corruption, label noise, or long-tailed class distributions remains largely under-explored. Our experiments reveal that no single approach excels uniformly across these diverse scenarios, as depicted in Figure~\ref{figure::teaser}. In this work, we propose a unified model designed to effectively address all these challenges.

In our pursuit to enhance uncertainty estimation, we start by examining the combined impact of several pre-existing methods, leading to the discovery of an integrated approach that significantly refines this estimation. We classify these methods based on their function in the model training process: regularization, classifier and optimization.
For regularization, we utilize techniques such as RegMixup regularization~\cite{pinto2022using}, correctness ranking loss (CRL)~\cite{moon2020confidence} and cosine similarity classifier (CSC)~\cite{gidaris2018dynamic,hu2020empirical}, which can help in increasing entropy for challenging samples. In the realm of optimization, we incorporate Sharpness-Aware Minimization (SAM)~\cite{foret2020sharpness} and Stochastic Weight Averaging (SWA)~\cite{izmailov2018averaging}, as recommended by FMFP~\cite{zhu2022rethinking}, to ensure that the model can converge towards flatter minima. 
The synergistic integration of these diverse techniques culminates in our novel approach, which we name SURE. This method harnesses the strengths of each individual component, resulting in a more robust and reliable model.

In the evaluation of SURE, we first focus on failure prediction, a pivotal task for evaluating uncertainty estimation. Our evaluations reveal that SURE consistently outperforms models deploying individual technique. This superior performance is evident across various datasets such as CIFAR-10~\cite{krizhevsky2009learning}, CIFAR-100~\cite{krizhevsky2009learning}, Tiny-ImageNet \cite{le2015tiny} and also across various model architectures, namely ResNet~\cite{he2016deep}, VGG~\cite{simonyan2014very}, DenseNet~\cite{huang2017densely}, WideResNet~\cite{zagoruyko2016wide} and DeiT \cite{touvron2021training}. Notably, SURE even surpasses OpenMix~\cite{zhu2023openmix}, a method that leverages additional OOD data.
By applying SURE directly to real-world scenarios, without or with minimal task-specific adjustments, we further witness its effectiveness in bringing robustness to the models. Specifically, the real-world challenges include data corruption in CIFAR10-C~\cite{hendrycks2019benchmarking}, label noise in Animal-10N~\cite{song2019selfie} and Food-101N~\cite{lee2018cleannet}, and skewed class distribution in CIFAR-LT~\cite{cui2019class}. In these contexts, SURE achieves results that are either superior to or on par with the latest specialized methods. A standout achievement is observed on Food-101N, where SURE attains an impressive accuracy of 88.0\%, significantly surpassing the previous state-of-the-art method, Jigsaw-ViT \cite{chen2023jigsaw}, which achieved accuracy of 86.7\% by using extra training data to pre-train the model. This demonstrates SURE's remarkable capability in handling complex real-world data challenges.

The main contributions of this paper are summarized as follows: 
\begin{itemize}
\item We reveal that existing methods do not uniformly excel in various real-world challenges. This analysis underlines the need for more reliable and robust approaches to handle the complexities of real-world data.
\item We propose a novel approach, named SURE, for robust uncertainty estimation, inspired by the synergistic effect achieved by combining multiple techniques, across model regularization, classifier and optimization. Models trained under our SURE approach consistently achieve better performance in failure prediction than models that deploy individual technique, across various datasets and model architectures. 
\item When applied directly to real-world scenarios, SURE consistently shows performance at least comparable to state-of-the-art specialized methods.
\end{itemize}

\section{Related work}
\paragraph{Uncertainty estimation}
Quantifying uncertainty for DNN outputs can improve the interpretability and trustworthiness of the predictions and serve various downstream tasks, such as model calibration~\cite{guo2017calibration}, OOD detection~\cite{liang2017enhancing,hendrycks2018deep}, failure prediction~\cite{corbiere2021confidence,hendrycks2016baseline}, etc. MSP~\cite{hendrycks2016baseline}, Entropy~\cite{steinhardt2016unsupervised}, and Energy~\cite{liu2020energy} provide uncertainty estimates for outputs using the information provided by the DNN itself. Modifying the architecture and optimization of the DNN can further improve the performance of these measures on downstream tasks, \textit{i.e.}, attaining robust and reliable uncertainty estimates. To balance the sensitivity and smoothness of the DNN and achieve robust uncertainty estimation, DDU~\cite{Mukhoti_2023_CVPR} applies spectral normalization layers~\cite{miyato2018spectral} to encourage bi-Lipschitzness and LDU~\cite{franchi2022latent} introduces distinction maximization layer and an uncertainty estimation head to the DNN. Yet, they all lead to increased training parameters, and a predefined input image size is needed for the former, which also lacks scalability.  
A simpler adjustment to DNN architecture introduced by OVADM~\cite{padhy2020revisiting} improves OOD detection performance, which replaces the output layer with an $\ell_2$ distance-based layer and uses a one-vs-all loss for training. In terms of optimization, in addition to FMFP~\cite{zhu2022rethinking} mentioned in the previous section, Qu et al.~\cite{qu2023towards} use meta-learning to achieve flat minima yet apply to the auxiliary uncertainty estimators.
Methods based on data augmentation, such as Mixup~\cite{zhang2017mixup}, RegMixup~\cite{pinto2022using} and OpenMix~\cite{zhu2023openmix}, apply regularization when training the model, resulting in dependable uncertainty estimates,
while ensuring classification accuracy. This work selects and integrates these methods and obtains a scalable solution to improve classification accuracy with more reliable uncertainty estimates.

\begin{figure*}
\centering
\includegraphics[width=0.9\textwidth]{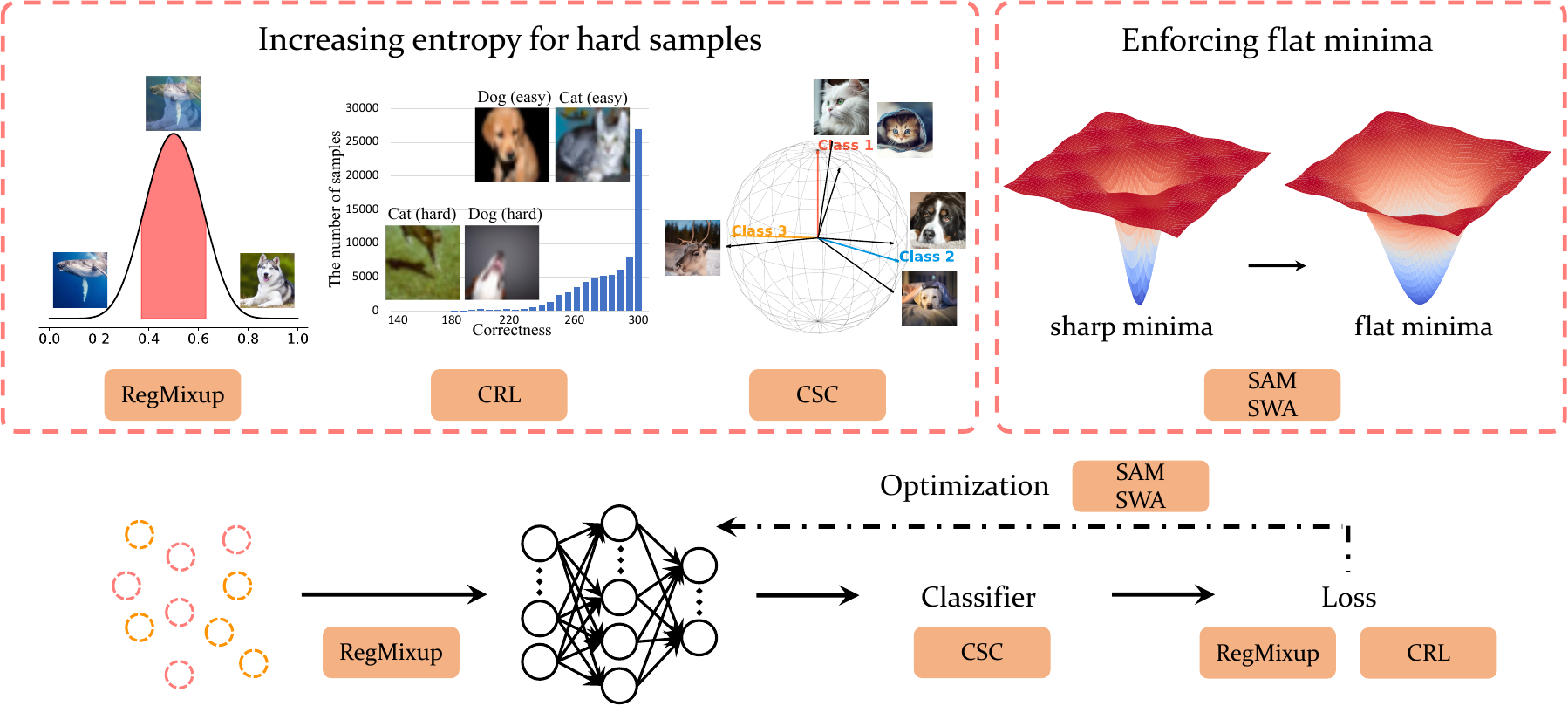}
\caption{\textbf{Overview of recipes.} Our proposed approach SURE contains two aspects: increasing entropy for hard samples and enforcing flat minima during optimization. We incorporate RegMixup~\cite{pinto2022using} loss and correctness ranking loss (CRL)~\cite{moon2020confidence} as our loss function and employ cosine similarity classifier (CSC) ~\cite{gidaris2018dynamic,hu2020empirical} as our classifier to increase entropy for hard samples. As in optimization, we leverage Sharpness-Aware Minimization (SAM)~\cite{foret2020sharpness} and Stochastic Weight Averaging (SWA)~\cite{izmailov2018averaging} to find flat minima.} 
\label{figure::overview method}
\end{figure*}

\paragraph{Learning with noisy labels}%
This task aims to perform learning while noisy annotated data is presented in the training set. 
Mainstream solutions include: 
\textit{i)} label correction, which aims at revising possibly wrong labels with more consistent substitutes~\cite{tanaka2018joint,yi2019probabilistic,song2019selfie,zhang2021learning};
\textit{ii)} semi-supervised learning, which trains networks in a semi-supervised manner with only the clean labels used \cite{ding2018semi,li2020dividemix,kong2019recycling,berthelot2019mixmatch}; 
\textit{iii)} sample re-weighting, which assigns more weights to possibly clean samples~\cite{chang2017active,han2018co,jiang2018mentornet,malach2017decoupling,yu2019does,wei2020combating,pleiss2020identifying,chen2022compressing,feng2022ssr}; 
\textit{iv)} over-fitting prevention, which prevents networks from over-fitting on noisy training data so as to have better generalization on clean test set~\cite{ma2018dimensionality,liu2020early,nishi2021augmentation,zheltonozhskii2022contrast,jiang2023dynamic,chen2022compressing}. Specifically, WarPI~\cite{sun2022learning} and~\cite{jiang2023dynamic} are based on meta-learning framework, which propose adaptively rectifying the training procedure for the classification network.
SSR+~\cite{feng2022ssr} designs a sample selection and relabelling based on a non-parametric KNN classifier and a parametric classifier.

\paragraph{Long-tailed classification}%
In addressing the long-tailed classification challenge, various strategies have been proposed. 
BBN \cite{zhou2020bbn} utilizes a dual-branch network to balance learning between different class frequencies, while SSP \cite{yang2020rethinking} leverages self-supervised learning and semi-supervised learning for contrastive learning in long-tailed distribution. 
LDAM-DRW \cite{cao2019learning} introduces logit compensation to handle class frequency imbalance. 
Hybrid-SC \cite{wang2021contrastive} proposes a two-branch network for supervised contrastive learning and reducing classifier bias. 
BCL \cite{zhu2022balanced} develops a balanced contrastive loss, ensuring that all classes are optimized for a regular simplex configuration that yields a balanced feature space. 
Recently, GLMC \cite{du2023global} proposes a new paradigm that contains a global and local mixture consistency loss to improve the robustness of the feature extractor, and a cumulative head-tail soft label re-weighted loss to mitigate the head class bias problem.

In this work, we show that by simply applying the uncertainty score provided by DNNs trained using SURE to the re-weighting training strategy, which is commonly used in the community of long-tailed classification \cite{alshammari2022long, cao2019learning, kang2019decoupling, zhou2020bbn, tang2020long}, the classification performance on imbalanced data can on par with the previous SOTAs.

\section{Methods}

As illustrated in Figure~\ref{figure::overview method}, 
our proposed approach SURE aims to train 
reliable and robust DNNs through two aspects: 
\textit{i)} increasing entropy for hard samples; \textit{ii)} enforcing flat minima during optimization.
In the following, we denote the dataset by $\{(\imagei, \targeti)\}_{i=1}^N$ where $\imagei$ is the input image, $\targeti$ is its ground-truth label and $N$ is the number of samples.

The recipes in SURE for increasing entropy for hard samples consist of three components: the RegMixup regularization~\cite{pinto2022using} denoted as $\lossmixup$, 
the correctness ranking loss $\losscrl$ which serves to regularize the class probabilities by aligning the confidence with the ordinal ranking of correctness, and the cosine similarity classifier (CSC). 
These recipes are employed collectively to optimize the objective, which includes a task-specific loss, \textit{e.g.}, the cross-entropy loss for classification, denoted as $\losscls$, in addition to the RegMixup regularization $\lossmixup$, and the confidence-aware regularization $\losscrl$ based on the historical correctness information gathered during training.
The recipes for enforcing flat minima lie in leveraging Sharpness-Aware Minimization (SAM)~\cite{foret2020sharpness} and Stochastic Weight Averaging (SWA)~\cite{izmailov2018averaging} during optimization.

This section is organized as follows: Section~\ref{section::goal} illustrates our objective function and CSC to increase entropy for hard samples. Section~\ref{section::optim} introduces the flat minima-enforced techniques. Implementation details are provided in Section~\ref{sec:details}.

\subsection{Increasing entropy for hard samples}
\label{section::goal}
\paragraph{Total loss}
As described above, the objective function of SURE is composed of three components,
which is expressed as:
\begin{equation} \label{eq:losstot}
    \losstot = \losscls + \weightmixup\lossmixup + \weightcrl\losscrl,
\end{equation}
where $\weightmixup$ and $\weightcrl$ denote hyper-parameters to balance the contribution of each loss component to the total loss. The impact of $\weightmixup$ and $\weightcrl$ is studied in the Apendix \ref{section::lambda}.

\paragraph{RegMixup regularization $\lossmixup$}
Mixup~\cite{zhang2017mixup} is a widely used data augmentation for image classification. Given two input-target pairs ($\imagei$,~$\targeti$) and ($\imagej$,~$\targetj$), we obtain an augmented sample ($\imagemixup$,~$\targetmixup$) by linearly interpolating between them: 
\begin{equation} \label{eqn:mixup}
    \imagemixup = \mixcoef \imagei + (1 - \mixcoef) \imagej, \quad \targetmixup = \mixcoef \targeti + (1 - \mixcoef)
    \targetj,
\end{equation}
\noindent where $\mixcoef$ denotes the mixing coefficient, following a \text{Beta} distribution: 
\begin{equation} \label{eqn:beta}
    \mixcoef \sim \text{Beta}(\betadist,  \betadist), \quad \betadist \in (0, \infty).
\end{equation}
The RegMixup regularization $\lossmixup$ consists of fitting the model additionally on the augmented samples ($\imagemixup$,~$\targetmixup$) : 
\begin{equation} \label{eqn:clsmixup}
    \lossmixup (\imagemixup, \targetmixup) = 
    \losscls(\imagemixup, \targetmixup),
\end{equation}
with $\beta=10$ leading to a heavy mixing of two samples with high probability.

Similar to RegMixup~\cite{pinto2022using}, we incorporate $\lossmixup$ as an additional regularizer alongside the original cross-entropy loss on $(\imagei,\targeti)$, \textit{i.e.}, $\losscls$ in~\eqref{eq:losstot}. A high value of $\beta$ results in a heavy mixing of samples, prompting the model to exhibit high entropy on heavily interpolated samples, which can be regarded as challenging examples.

\paragraph{Correctness ranking loss $\losscrl$}
The correctness ranking loss~\cite{moon2020confidence} encourages the DNN to align the model's confidence with the ordinal ranking of historical correctness information gathered during training. Specifically, for two input images $\imagei$ and $\imagej$, $\losscrl$ is defined as:
\begin{equation} \label{eqn:crl}
    \losscrl(\imagei, \imagej) = \max(0, |c_i - c_j| -\sgn(c_i - c_j)(\scorei - \scorej)),
\end{equation}
where \(c_i\) and \(c_j\) represent the proportions of correct prediction events for $\imagei$ and $\imagej$ during the training, $\scorei$ and $\scorej$ denote the confidence score for $\imagei$ and $\imagej$, which are the softmax scores in this work, $\sgn$ denotes the sign function,
$\losscrl$ aims to align the confidence score to the correctness statistics. Hard samples, which are less likely to be correctly predicted during training, are encouraged to have lower confidence and thus, higher entropy.

\paragraph{Cosine Similarity Classifier (CSC)}
CSC has shown to be effective on few-shot classification~\cite{gidaris2018dynamic,hu2020empirical} by simply replacing the last linear layer with a cosine classifier. %
For the image $\imagei$, we denote the classification logits for $\imagei$ belonging to class \( k \) as $\scorei^k$, which is defined as follows: 
\begin{equation} \label{eqn:cos}
    \scorei^k = \tau \cdot \cos(\network_\netweight(\imagei), w^k) = \tau \cdot \frac{\network_\netweight(\imagei)}{\|\network_\netweight(\imagei)\|}_2 \cdot \frac{w^k}{\|w^k\|_2},
\end{equation}
where \( \tau \) is the temperature hyper-parameter,
$\network_\netweight$ is a DNN parameterized with $\netweight$, used to extract features of input images, $w^k$ representing the $k$-th class prototype, denotes the weight of the $k$-th class. 

CSC encourages the classifier to focus on the directional alignment between the feature vector extracted from the input image and the class prototype vector, rather than the dot product. This makes it conceptually distinct from the traditional linear classifier, where magnitude plays a significant role.
A key benefit of using CSC is its ability to handle hard samples better. CSC views hard samples as equidistant in angle to several class prototypes, leading to more effective interpretation and potentially higher entropy than the traditional linear classifier that uses the dot product.

\subsection{Flat minima-enforced optimization}
We jointly employ Sharpness-Aware Minimization (SAM) \cite{foret2020sharpness} and Stochastic Weight Averaging (SWA) \cite{izmailov2018averaging} to enhance flat minima. Note that these two techniques are also jointly used in~\cite{zhu2022rethinking} to improve uncertainty estimation.
\label{section::optim}
\paragraph{Sharpness-Aware Minimization (SAM)}
SAM \cite{foret2020sharpness,chen2023does} is an optimization method that enhances model generalization by seeking parameters lying in flat neighborhoods such that the DNN has a uniformly small loss. For our objective function $\losstot$ and DNN parameters $\netweight$, the SAM optimizer seeks $\netweight$ satisfying:
\begin{equation} \label{eqn:sam}
    \min_{\netweight} \max_{\|\epsilon\|_2 \leq \rho} \losstot( \netweight+ \epsilon),
\end{equation}

\noindent where $\epsilon$ is a perturbation vector and $\rho$ is the neighborhood size within which we seek to minimize the sharpness of the loss. The SAM algorithm proceeds by alternating between finding the worst-case perturbation $\epsilon$ that maximizes the loss within the \( \ell_2 \)-norm ball of radius $\rho$, and updating the model parameters $\netweight$ to minimize this perturbed loss.  %

\begin{table*}[!t]
    \begin{center}
		\resizebox{2\columnwidth}{!}{
			\begin{tabular}{|l|c|c|c|c|c|c|c|c|c|c|c|c|c|c|c|}
				\hline \hline 
				\multirow{2}{*}{Backbones}&\multirow{2}{*}{Methods} &\multicolumn{4}{c|}{CIFAR-10 \cite{krizhevsky2009learning}} &\multicolumn{4}{c|}{CIFAR-100 \cite{krizhevsky2009learning} } &\multicolumn{4}{c|}{Tiny-ImageNet \cite{le2015tiny}}    \\ 
                    
                        & &Acc.  $\uparrow$  &AURC $\downarrow$ &AUROC $\uparrow$ & FPR95 $\downarrow$
                        &Acc.  $\uparrow$  &AURC $\downarrow$ &AUROC $\uparrow$ & FPR95 $\downarrow$
                        &Acc.  $\uparrow$  &AURC $\downarrow$ &AUROC $\uparrow$ & FPR95 $\downarrow$
                        \\ \hline 
			         
				    & MSP \cite{hendrycks2016baseline} & 94.89±0.20   & 6.78±0.33 & 92.20±0.55& 38.73±2.89 & 75.87±0.31 & 69.44±2.11 & 87.00±0.21  & 60.73±1.16   & 63.39±0.59 &136.50±1.08 & 85.62±0.35 & 63.99±0.64
\\
				      & RegMixup \cite{pinto2022using}& 95.69±0.13   & 4.74±0.27 &92.96±0.29 &34.26±1.98 & 77.90±0.37 &59.23±1.65 & 87.61±0.13  & 58.65±0.43 & 66.36±0.43
                        & 115.08±1.98
                        & 86.53±0.27
                        & 62.54±0.43  \\
                        & CRL \cite{moon2020confidence} & 94.85±0.10   & 5.09±0.28 &93.64±0.48& 35.33±1.73& 76.42±0.21 & 62.78±0.21 & 88.07±0.17  & 59.02±0.39   & 65.50±0.03  & 117.46±0.56 & 87.01±0.13 & 61.15±0.07 \\
\bf ResNet-18 \cite{he2016deep} & SAM \cite{foret2020sharpness} & 95.30±0.25   & 3.97±0.33 & 	94.53±0.31& 31.13±3.62&76.60±0.21 & 62.97±1.02 &87.72±0.10  & 59.35±0.87 &64.95±0.21
 & 120.04±2.11
 &87.19±0.57
  & \bf 59.98±0.55\\
                        & SWA \cite{izmailov2018averaging}& 95.38±0.09   & 4.00±0.21 &94.40±0.50&35.70±1.44 & 77.65±0.19 & 55.87±0.32 & 88.55±0.25 & 60.43±1.90    &68.09±0.19 &102.11±0.51
& 87.27±0.15 & 60.63±1.38
 \\
                        & FMFP \cite{zhu2022rethinking}& 95.60±0.09   & 3.56±0.06 &	94.74±0.10&33.49±0.33& 77.82±0.08 & 55.03±0.52 & 88.59±0.07  &  59.79±0.31  & 68.18±0.42
   & 100.93±2.12
  &87.45±0.05
  & 60.18±1.26\\		
				    &  \bf SURE& \bf 96.14±0.16 & \bf 2.97±0.13& \bf 95.08±0.04 & \bf 28.64±0.66
 &  \bf 80.49±0.18& \bf 45.81±0.15& \bf 88.73±0.24  & \bf 58.91±0.58   &\bf 69.55±0.10	& \bf 93.46±0.82 & \bf 87.67±0.12 & 	60.13±0.32\\
		              \hline \hline
				    & MSP \cite{hendrycks2016baseline}  &  93.30±0.21  &  10.41±0.33 & 	90.71±0.04& 44.66±1.81&72.43±0.42 & 91.40±1.95 & 85.69±0.90  & 64.41±1.66    & 59.52±0.62 & 156.45±2.51 & 86.33±0.63 & 63.79±0.95 \\
				      & RegMixup \cite{pinto2022using} & 94.11±0.28    & 9.89±0.81 &	89.90±0.26&39.93±1.58& 73.51±0.18 & 85.98±1.05 & 86.35±0.32  & 61.70±1.83   & 63.04±0.57 & 146.72±2.59 & 85.60±0.39 & \bf 59.00±1.27\\
                        & CRL \cite{moon2020confidence} & 93.42±0.09   & 7.61±0.44 &	92.88±0.56&39.66±2.83& 72.63±0.27 & 80.94±0.47& 87.37±0.28   &61.96±0.77  & 60.20±0.36 & 146.76±1.42 & \bf 87.42±0.28 & 59.26±1.44\\
\bf VGG \cite{simonyan2014very} & SAM \cite{foret2020sharpness}&  94.11±0.06  & 5.97±0.08&	93.68±0.13& 37.21±2.92 &73.33±0.36  & 77.44±0.75 & 87.42±0.33  & 63.19±0.58 & 61.24±0.07 & 142.54±1.04 & 86.82±0.25 & 62.93±1.12\\
                        & SWA \cite{izmailov2018averaging}& 93.76±0.25   & 6.64±0.24 &93.43±0.16&40.44±1.27& 73.98±0.16 & 74.23±0.58 & 87.30±0.14  & 62.89±1.80	   & 62.48±0.19  & 137.01±0.71 & 86.29±0.16 & 62.15±1.64 \\
                        & FMFP \cite{zhu2022rethinking}& 94.26±0.23   & 5.89±0.16 &93.46±0.26&40.67±3.14	& 74.77±0.31 & 70.07±1.26& 87.58±0.19 &60.98±1.16& 62.95±0.16 & 134.04±1.42 & 86.36±0.12 & 61.71±1.08 \\		
				    & \bf \bf SURE& \bf 95.00±0.11   & \bf 4.98±0.24 & \bf 93.79±0.62& \bf35.92±2.95	& \bf 76.51±0.07 & \bf 65.25±0.17& \bf87.59±0.07  & \bf 60.27±0.60   & \bf 63.75±0.11
 &  \bf 131.40±0.28
  & 86.12±0.19
  & 63.04±1.05\\
		            \hline \hline
                        & MSP \cite{hendrycks2016baseline} &  94.72±0.23  & 5.94±0.23 &	93.00±0.45&37.00±0.31& 75.14±0.07 & 74.68±0.32& 86.22±0.22  & 62.79±0.80   & 57.90±0.25 & 180.08±2.52 & 83.65±0.29 & 68.61±0.37  \\
				      & RegMixup \cite{pinto2022using} & 95.13±0.22   & 6.03±0.50 &	92.20±0.80&38.63±1.63& 77.29±0.16 & 63.96±1.15 & 86.57±0.07  &  63.76±1.10  & 61.96±0.09 & 147.22±1.57 & \bf 84.91±0.17 & 65.92±0.40 \\
                        & CRL \cite{moon2020confidence}&  94.79±0.02  & 5.58±0.42 &	93.22±0.61&37.34±2.73& 76.09±0.06 & 65.96±0.62& 87.41±0.11  & 60.67±0.72    &58.80±0.56 & 169.44±3.74 & 84.49±0.04 & 66.05±0.60\\
\bf DenseNet \cite{huang2017densely}& SAM \cite{foret2020sharpness}&  95.31±0.10  & 4.25±0.17 &94.15±0.46&33.33±1.27 &78.17±0.26  &  57.20±0.73 & 86.99±0.23  & 61.42±0.74  & 60.49±0.31 & 158.94±3.86 & 84.39±0.57 & 66.51±1.85 \\
                        & SWA \cite{izmailov2018averaging}& 94.86±0.09 & 4.65±0.18 &	94.27±0.27&35.78±4.61& 78.17±0.26 & 57.20±0.73 & 87.23±0.22  & 63.33±0.63	   & 60.74±0.46 & 159.68±3.12  & 83.83±0.07 & 68.03±0.75 \\
                        & FMFP \cite{zhu2022rethinking}& 95.07±0.15 & 4.11±0.19 &94.74±0.06&34.67±0.48	& 78.33±0.40  & 54.88±1.62 & 87.92±0.46  & 60.52±1.12   & 61.18±0.72 & 154.98±3.72 & 84.29±0.26 & 66.66±1.21\\	
                        & OpenMix \cite{zhu2023openmix}$^\S$ & 95.51±0.23   & 4.68±0.72  &93.57±0.81 &33.57±3.70& 78.97±0.31
  & 53.83±0.93 & 87.45±0.18 & 62.22±1.15    & - & - & - & -\\
				    & \bf SURE& \bf 95.57±0.06   &  \bf 3.51±0.09 & \bf94.91±0.25& \bf29.52±0.56& \bf 80.02±0.13 & \bf 46.69±0.59 & \bf88.78±0.26  & \bf58.37±0.39   &  \bf 62.61±0.18	& \bf 142.59±2.16&	84.31±0.42	& \bf 65.39±2.12 \\
		              \hline \hline
                        & MSP \cite{hendrycks2016baseline}& 95.71±0.17   &  5.90±0.89  &	92.19±0.82&35.95±3.75&  79.15±0.19 &  53.02±0.89& 88.21±0.06  & 59.46±1.23  & 67.52±0.18  & 107.97±0.80 &86.78±0.20&61.68±0.99\\
				      & RegMixup \cite{pinto2022using}&  97.03±0.04  & 3.47±0.26 &	93.10±0.56&26.16±1.17& 82.14±0.47 & 47.01±2.12 & 87.70±0.17  & 55.24±1.19   &69.63±0.09  & 95.96±0.21 &87.38±0.21   & 59.09±0.75\\
                        & CRL \cite{moon2020confidence} & 95.87±0.08   & 3.85±0.20 &94.10±0.06&32.73±1.22	& 80.10±0.28 & 47.99±1.08& 88.43±0.34  & 59.44±1.45   & 69.00±0.22 & 97.46±0.90 & 87.42±0.23 & 61.02±1.71\\
\bf WRNet \cite{zagoruyko2016wide}& SAM \cite{foret2020sharpness}&  96.47±0.11  & 2.91±0.38  &	94.79±0.29&28.05±1.56& 80.67±0.31  & 44.93±0.87&89.01±0.31   & 56.60±1.30   & 69.86±0.37 & 93.66±2.03 & 87.49±0.30 & 60.44±1.19 \\
                        & SWA \cite{izmailov2018averaging}& 94.86±0.09   & 4.65±0.18 &94.27±0.27&35.78±4.61& 81.31±0.33  &  41.15±0.89& 89.39±0.16  & 57.57±1.97   & 71.27±0.16 & 84.97±0.12 & 87.71±0.26 & 60.00±2.42\\
                        & FMFP \cite{zhu2022rethinking}&  96.47±0.12   & 2.33±0.08  &95.73±0.01&26.68±2.62& 81.66±0.12  &  39.60±0.15& 89.51±0.10  & 56.41±1.44   & 71.62±0.04 & 83.04±0.16 & 87.78±0.03 & 60.09±0.83\\	
                        & OpenMix \cite{zhu2023openmix}$^\S$ & \bf 97.16±0.10   & 2.32±0.15  &94.81±0.34 & 22.08±1.86&  82.63±0.06
  & 39.61±0.54 & 89.06±0.11 & 55.00±1.29     & - & - & - & -\\
				    &\bf SURE& 97.02±0.20 & \bf 1.79±0.16& \bf96.18±0.01& \bf19.53±1.23&  \bf 83.71±0.10  &  \bf \bf32.10±0.28& 	\bf90.33±0.18 &  \bf54.34±0.29  &\bf 73.34±0.36 & \bf 74.11±0.97
 & \bf 88.23±0.31  & \bf 58.17±1.50\\
		            \hline \hline
              & MSP \cite{hendrycks2016baseline}&98.28±0.08&0.97±0.02&95.76±0.28&	20.47±5.38
&89.71±0.03	&17.66±0.56	&90.40±0.25	&50.99±0.61 		      
                        &-&-&	-	&-
\\
				      & RegMixup \cite{pinto2022using}&98.90±0.04&0.89±0.05&	94.30±0.25&	24.98±3.87
&90.79±0.11&15.38±0.51&	90.34±0.33	&52.01±1.76
&-	&-&	-&-
\\\
                        & CRL \cite{moon2020confidence} &98.27±0.04&0.99±0.11&	95.85±0.44&	19.65±2.51
&89.74±0.16&17.61±0.71&	90.30±0.18	&51.58±0.23
&-&-&-&-
\\\
\bf DeiT-B $^\star$ \cite{touvron2021training} & SAM \cite{foret2020sharpness}&98.62±0.10&0.58±0.09&96.89±0.34&\bf 15.74±1.71
&90.43±0.17&15.29±0.19&	90.75±0.15&	50.02±1.52
&-&-&-&-
\\\
                        & SWA \cite{izmailov2018averaging}&98.44±0.07&0.82±0.03	&96.11±0.20	&17.78±3.23
&90.17±0.34&	15.37±0.44	&90.86±0.38	&50.64±3.37
&-&	-	&-&	-
\\\
                        & FMFP \cite{zhu2022rethinking}&98.76±0.02& \bf 0.46±0.02&	\bf 97.15±0.16&	16.17±0.55
&90.53±0.13&	14.30±0.18	&\bf 91.15±0.32&	51.90±1.50
& - &	- & - & -
\\\         
				     & \bf SURE& \bf 98.92±0.07	&0.86±0.08	&94.37±0.69&	27.52±3.11
&\bf 91.18±0.01	& \bf 13.79±0.29	&90.85±0.05	& \bf 48.81±0.39
&  - & - & - & -
\\	
         \hline \hline	
		    \end{tabular}
        }
            \footnotesize{\\$^\S$ reports the results given by models training on extra outliers and all the training data on CIFAR10 \cite{krizhevsky2009learning} CIFAR100 \cite{krizhevsky2009learning}
            \\$^\star$ reports the results given by finetuning ImageNet \cite{deng2009imagenet}
pre-trained DeiT-B \cite{touvron2021training} for 50 epochs}
            \vspace{-2mm}
            \caption{\textbf{Comparison of the performance of failure prediction on CIFAR10 \cite{krizhevsky2009learning}, CIFAR100 \cite{krizhevsky2009learning} and Tiny-ImageNet \cite{le2015tiny}.} We keep 10\% training data as the validation set to select the best model. The means and standard deviations over \textit{three} runs are reported. $\downarrow$ and $\uparrow$ indicate that lower and higher values are better respectively. AURC \cite{geifman2018bias} values are multiplied by $10^{3}$, and all remaining values are in percentage. } 
		\label{tab::main_results}
	\end{center}
\end{table*}

\paragraph{Stochastic Weight Averaging (SWA)}
SWA is introduced in \cite{izmailov2018averaging}, which improves the generalization of DNNs by averaging model weights over the course of training. The process begins with a standard training phase, after which SWA starts by averaging the weights at each subsequent epoch. The mathematical representation of the SWA weight update is given by:
\begin{equation}
    \netweight_{\text{SWA}} = \frac{1}{T} \sum\nolimits_{t=1}^{T} \netweight_t,
\end{equation}
where $\netweight_t$ 
represents the model weights at epoch $t$ 
, and $T$ is the total number of epochs during which SWA is applied.

\subsection{Implementation details}
\label{sec:details}

Following~\cite{zhu2022rethinking}, our models are trained using SAM~\cite{foret2020sharpness} with stochastic gradient descent (SGD) as the base optimizer with a momentum of 0.9, starting with an initial learning rate of 0.1 and a weight decay of 5e-4, over 200 epochs with a batch size of 128. 
We employ a cosine annealing learning rate schedule and set the SWA \cite{izmailov2018averaging} start epoch to 120 and a SWA-specific learning rate of 0.05, to enhance the training effectiveness and model robustness. We set $\beta$ = 10 in~\eqref{eqn:beta} for the Mixup data augmentation, which is following~\cite{pinto2022using}. All hyper-parameters, including $\weightmixup$, $\weightcrl$, and $\tau$, are selected on the validation set. An ablation study of $\weightmixup$in~\eqref{eq:losstot}, $\weightcrl$ in~\eqref{eq:losstot}, and $\tau$ in~\eqref{eqn:cos} are provided in Appendix \ref{section::lambda}, \ref{section::tau}.
In terms of finetuning DeiT-Base \cite{touvron2021training} with the ImageNet \cite{deng2009imagenet} pre-trained model, we set the learning rate at 0.01 with a weight decay of 5e-5 over 50 epochs and start SWA start epoch to 1 and a SWA-specific learning rate of 0.004.

\section{Experiments}
In this section, we evaluate the performance of SURE in failure prediction and further explore SURE's ability in tackling real-world challenges, including long-tailed classification, learning with noisy labels, and generalization in corrupted image scenarios. We first introduce the datasets used in our experiments and outline the key metrics  in Section~\ref{sec::datasets}. 
Then, we present results on failure prediction in Section~\ref{sec::failure prediction}. Results on long-tailed classification are presented in Sections~\ref{sec::longtail}. In Section~\ref{sec::labelnoise}, we present results for learning with noisy labels. Performances on corrupted images are provided in Section~\ref{sec::corruption}. Finally, we present analysis in Section~\ref{section::ablation}.

\subsection{Datasets and evaluation metrics}
\label{sec::datasets}
\paragraph{CIFAR10, CIFAR100 and Tiny-ImageNet} We use CIFAR~\cite{krizhevsky2009learning} and Tiny-ImageNet \cite{le2015tiny} to evaluate failure prediction. CIFAR datasets are commonly used in the community~\cite{zhu2022rethinking,pinto2022using,zhu2023openmix} and we use Tiny-ImageNet \cite{le2015tiny} as a larger dataset to evaluate the effectiveness and robustness of our proposed method. The CIFAR10 dataset contains 60,000 color images with a resolution of 32$\times$32, divided into 10 classes, each holding 5,000 training images and 1,000 testing images. The CIFAR100 dataset follows a similar structure, but with 100 classes. Each class contains 500 training samples and 100 testing samples.
Tiny-ImageNet \cite{le2015tiny} contains 100,000 images
of 200 classes downsized to 64$\times$64 colored images which are a subset of the
ImageNet dataset \cite{deng2009imagenet}. Each class has 500 training images. 50 images are collected for testing. Note that for all our experiments, we keep 10\% of the training set as our validation set.  
We report the means and standard deviations over \textit{three} runs.

\paragraph{Long-Tailed CIFAR: CIFAR10-LT and CIFAR100-LT} We use CIFAR10-LT and CIFAR100-LT~\cite{cui2019class} to evaluate long-tailed classification. Note that these datasets are widely used as evaluation datasets in the community~\cite{alshammari2022long,zhu2022balanced,du2023global}. Following previous works~\cite{alshammari2022long,zhu2022balanced,du2023global}, the datasets are created by only keeping the number of training samples per class according to an exponential function $\tilde{N}_i = N_{i}\mu^{i} $ where $i$ is the class index, $N_{i}$
is the number of training images in the $i$-th class and $\mu \in (0, 1) $. The imbalanced factor $\text{IF}$ quantifies the level of distribution imbalance and determines $\mu$, which is defined by the ratio between the maximum and the minimum number of samples in a category. The test set remains unchanged.

\paragraph{Animal-10N and Food-101N}
Animal-10N~\cite{song2019selfie} and Food-101N~\cite{lee2018cleannet} are two real-world datasets where noisy labels are present in the training set. Animal-10N is a benchmark that contains 10 animal classes with confusing appearance. The training set size is 50,000, and the test set is 5,000. The estimated label noise ratio of the training set is 8\%. No data augmentation is applied so as to follow
the settings in~\cite{song2019selfie}. Food-101N contains 310,009 training images of different food recipes collected online and are classified into 101 classes. The training set is with an approximate noise ratio of 20\%. Following~\cite{lee2018cleannet}, the learned models should be evaluated on the test set of Food-101 with 25,250 clean labeled images.

\paragraph{CIFAR10-C}
To evaluate the model's robustness, we use CIFAR10-C dataset~\cite{hendrycks2019benchmarking}, which applies 15 common image corruptions, \textit{e.g.}, Gaussian noise, impulse noise, motion blur, frost, etc., to CIFAR10~\cite{krizhevsky2009learning} test set. Each type of corruption is characterized by five severity levels, as these corruptions can occur at different intensities.

\paragraph{Evaluation Metrics}
We report metrics that are commonly used in the community of failure prediction to assess the performance of our model, including Accuracy (Acc.), Area Under the Risk-Coverage Curve (AURC) \cite{geifman2018bias}, Area Under the Receiver Operating Characteristic Curve (AUROC) \cite{davis2006relationship}, False Positive Rate at 95\% True Positive Rate (FPR95). Specifically, we leverage AURC, which is complementary to Accuracy to measure the uncertainty of the model. AURC measures the area under the curve drawn by plotting the risk according to coverage. Given a confidence threshold, the coverage indicates the ratio of samples whose confidence estimates are higher than the confidence threshold, and the risk, also known as the selective risk \cite{geifman2017selective}, is an error rate computed by using those samples. A lower value of AURC implies a higher accuracy, and correct and erroneous predictions can be well-separable by a confidence threshold. The definitions of AUROC \cite{davis2006relationship} and FPR95 are detailed in Appendix \ref{section::auroc}.

\subsection{Failure prediction}
\label{sec::failure prediction}
We present results on failure prediction on CIFAR10~\cite{krizhevsky2009learning}, CIFAR100~\cite{krizhevsky2009learning} and Tiny-ImageNet~\cite{le2015tiny} in Table~\ref{tab::main_results}. Experiments are conducted with different backbones: ResNet18 \cite{he2016deep}, VGG16-BN \cite{simonyan2014very}, DenseNetBC \cite{huang2017densely}, WRNet28 \cite{zagoruyko2016wide} and DeiT~\cite{touvron2021training}. The architectures and datasets are commonly used in the community~\cite{zhu2022rethinking,pinto2022using,zhu2023openmix}. Note that to ensure the reliability of our model and maintain the rigor and fairness of our experiments, we split 10\% of the training data as a validation set for the selection of hyper-parameters and report the performance on the test set. All the experiments are repeated \textit{three} times and we report the mean and the standard deviation in the table. From Table~\ref{tab::main_results}, we can see that our SURE achieves significantly better performance on almost all the metrics than all the competitive approaches across different datasets and diverse architectures, which demonstrates the effectiveness and robustness of our proposed approaches. Note that even though the latest approach OpenMix \cite{zhu2023openmix} trains on all the training sets as well as additional outlier data, our SURE still maintains a significant performance gain without using any additional data.

\subsection{Long-tailed classification}
\label{sec::longtail}

\paragraph{Uncertainty-aware re-weighting} When the training data distribution is imbalanced, we find that the second stage uncertainty-aware re-weighting can consistently improve the performance. Note that the two-stage training strategy is commonly used in the community of long-tailed classification \cite{alshammari2022long, cao2019learning, kang2019decoupling, zhou2020bbn, tang2020long}. The key difference is that we use the uncertainty scores obtained from our first-stage training for re-weighting. Precisely, during the first epoch of re-weighting, we save the maximum softmax score for each sample in the training set to serve as the uncertainty score. By applying an exponential mapping to the uncertainty score. We re-weight the cross-entropy loss for each sample using $e^{-s_i}$ and normalize the weights of all samples in a
training batch such that they sum up to one. This re-weighting process is carried out over 50 epochs with a learning rate of 5e-3. Further ablation studies about variations of our re-weighting mapping are in Appendix \ref{section::re-weighting}.
\begin{table}[t]
    \begin{center}
		\resizebox{\columnwidth}{!}{
			\begin{tabular}{|l|c|c|c|c|c|c|c|}
				\hline \hline 
				\multirow{2}{*} {Methods} &\multicolumn{3}{c|}{CIFAR10-LT  \cite{cui2019class}} &\multicolumn{3}{c|}{CIFAR100-LT  \cite{cui2019class} } \\ 
                    
                    &\multicolumn{1}{p{1.5cm}}{\centering IF=100}   
                    &\multicolumn{1}{p{1.5cm}}{\centering IF=50} 
                    &\multicolumn{1}{p{1.5cm}|}{\centering IF=10}
                    &\multicolumn{1}{p{1.5cm}}{\centering IF=100}   
                    &\multicolumn{1}{p{1.5cm}}{\centering IF=50} 
                    &\multicolumn{1}{p{1.5cm}|}{\centering IF=10} \\ \hline
			        CE &70.40 & 74.80 &  86.40 & 38.30 & 43.90 & 55.70 \\
                        Mixup \cite{zhang2017mixup}&73.06 & 77.82 & 87.1 &39.54 &54.99 & 58.02   \\
				      CB-Focal \cite{cui2019class}& 74.57 & 79.27& 87.10& 39.60& 45.17& 57.99   \\
                        LDAM-DRW \cite{cao2019learning} & 77.03 & 81.03 & 88.16 & 42.04 & 46.62 & 58.71 \\
                        SSP \cite{yang2020rethinking}& 77.83 & 82.13 & 88.53 & 43.43 &  47.11&  58.91 \\
                        BBN \cite{zhou2020bbn} & 79.82 & 81.18 & 88.32 & 42.56 & 47.02 & 59.12 \\
                        Casual model \cite{tang2020long} & 80.60 &  83.60 & 88.50 & 44.10 & 50.30 &  59.60 \\
                        MetaSAug-LDAM \cite{li2021metasaug} & 80.66 & 84.34 & 89.68 & 48.01 & 52.27 & 61.28 \\
                        Hybrid-SC \cite{wang2021contrastive}& 81.40 & 85.36 & 91.12 & 46.72 & 51.87 &63.05 \\
                        ResLT \cite{cui2022reslt} & 82.40 & 85.17 &  89.70 & 48.21 & 52.71 & 62.01 \\
                        Dynamic Loss \cite{jiang2023dynamic} & 82.95 &  88.30 & 91.24 & 50.14 & 54.51 & 63.99 \\
                        BCL \cite{zhu2022balanced} & 84.32 & 87.24 & 91.12 & 51.93 & 56.59 & 64.87 \\
                        GLMC \cite{du2023global}& \bf 87.75  & 90.18 & 94.04 & 55.88 & 61.08 & 70.74 \\
                        \bf SURE &83.28 &87.72 & 93.73&51.60 & 58.57&  71.13  \\ \hline
                        GLMC + MaxNorm \cite{alshammari2022long} & 87.57  & \bf 90.22 & 94.03 & 57.11 & 62.32 & 72.33 \\
                        \bf SURE + re-weighting & 86.93 & \bf 90.22 & \bf 94.96 & \bf 57.34  & \bf 63.13 & \bf73.24 \\
                        \hline \hline
            
		    \end{tabular}}
      \vspace{-2mm}
      \caption{\textbf{Top-1 accuracy (\%) of ResNet32 \cite{he2016deep} on CIFAR10-LT and CIFAR100-LT \cite{cui2019class} with different imbalance factors [100, 50, 10].} SURE, enhanced with re-weighting, achieves comparable top-1 accuracy to the SOTA method GLMC \cite{du2023global} + MaxNorm \cite{alshammari2022long}.}
		\label{tab::longtail_results}
	\end{center}
\end{table}

\begin{table}[t]
    \begin{center}
	\resizebox{\columnwidth}{!}{
		\begin{tabular}{|c|ccccccc|c|}
            \hline \hline 
            Methods & \begin{tabular}[c]{@{}c@{}}CE\\~\cite{zhang2021learning}\end{tabular} 
            & \begin{tabular}[c]{@{}c@{}}SELFIE\\~\cite{song2019selfie}\end{tabular} 
            & \begin{tabular}[c]{@{}c@{}}PLC\\~\cite{zhang2021learning}\end{tabular} 
            & \begin{tabular}[c]{@{}c@{}}NCT\\~\cite{chen2022compressing}\end{tabular}  
            & \begin{tabular}[c]{@{}c@{}}Dynamic Loss\\~\cite{jiang2023dynamic}\end{tabular} 
            & \begin{tabular}[c]{@{}c@{}}SSR+\\~\cite{feng2022ssr}\end{tabular} & 
             \begin{tabular}[c]{@{}c@{}}Jigsaw-ViT $^\star$\\~\cite{chen2023jigsaw}\end{tabular} 
            & \textbf{SURE} 
            \\ \hline
            Acc.~(\%) 
            & 79.4 & 81.8 & 83.4 & 84.1 & 86.5 & 88.5 & 89.0 &\textbf{89.0} 
            \\ \hline \hline 
            \end{tabular}}
            \footnotesize{$^\star$ is with DeiT-S \cite{touvron2021training} and an extra self-supervised loss. The others are with VGG19-BN \cite{simonyan2014very}.}
            \vspace{-2mm}
		\caption{\textbf{Comparison of SOTA approaches on learning with noisy labels task on Animal-10N~\cite{song2019selfie} (noise ratio $\sim$8\%)}. Top-1 test accuracy (\%) is reported.}
        
		\label{tab::animal10n}
	\end{center}
\end{table}
\vspace{-1mm}
\begin{table}[t]
    \begin{center}
	\resizebox{\columnwidth}{!}{
		\begin{tabular}{|c|cccccccc|c|}
            \hline \hline 
            Methods 
            & \begin{tabular}[c]{@{}c@{}}CE\\~\cite{zhang2021learning}\end{tabular} 
            & \begin{tabular}[c]{@{}c@{}}CleanNet\\~\cite{lee2018cleannet}\end{tabular} 
            & \begin{tabular}[c]{@{}c@{}}MWNet\\~\cite{shu2019meta}\end{tabular} 
            & \begin{tabular}[c]{@{}c@{}}SMP\\~\cite{han2019deep}\end{tabular}  
            & \begin{tabular}[c]{@{}c@{}}NRank\\~\cite{sharma2020noiserank}\end{tabular} 
            & \begin{tabular}[c]{@{}c@{}}PLC\\~\cite{zhang2021learning}\end{tabular} 
            & \begin{tabular}[c]{@{}c@{}}WarPI\\~\cite{sun2022learning}\end{tabular}
            & 
             \begin{tabular}[c]{@{}c@{}}Jigsaw-ViT $^\star$\\~\cite{chen2023jigsaw}\end{tabular} 
            & \textbf{SURE} 
            \\ \hline
            Acc.~(\%) 
            & 81.7 & 83.5 & 84.7 & 85.1 & 85.2 & 85.3 & 85.9 & 86.7 & \textbf{88.0} 
            \\ \hline \hline 
            \end{tabular}}
            \footnotesize{$^\star$ is with DeiT-S \cite{touvron2021training} and an extra self-supervised loss. The others are with ResNet-50 \cite{he2016deep}.}
            \vspace{-2mm}
		\caption{\textbf{Comparison of SOTA approaches on learning with noisy labels task on Food-101N~\cite{lee2018cleannet} (noise ratio $\sim$20\%)}. Top-1 test accuracy (\%) is reported.}
		\label{tab::food101n}
	\end{center}
\end{table}

\begin{figure}[htbp]
\centering
\begin{subfigure}{0.46\textwidth}
    \centering
    \includegraphics[width=0.9\textwidth]{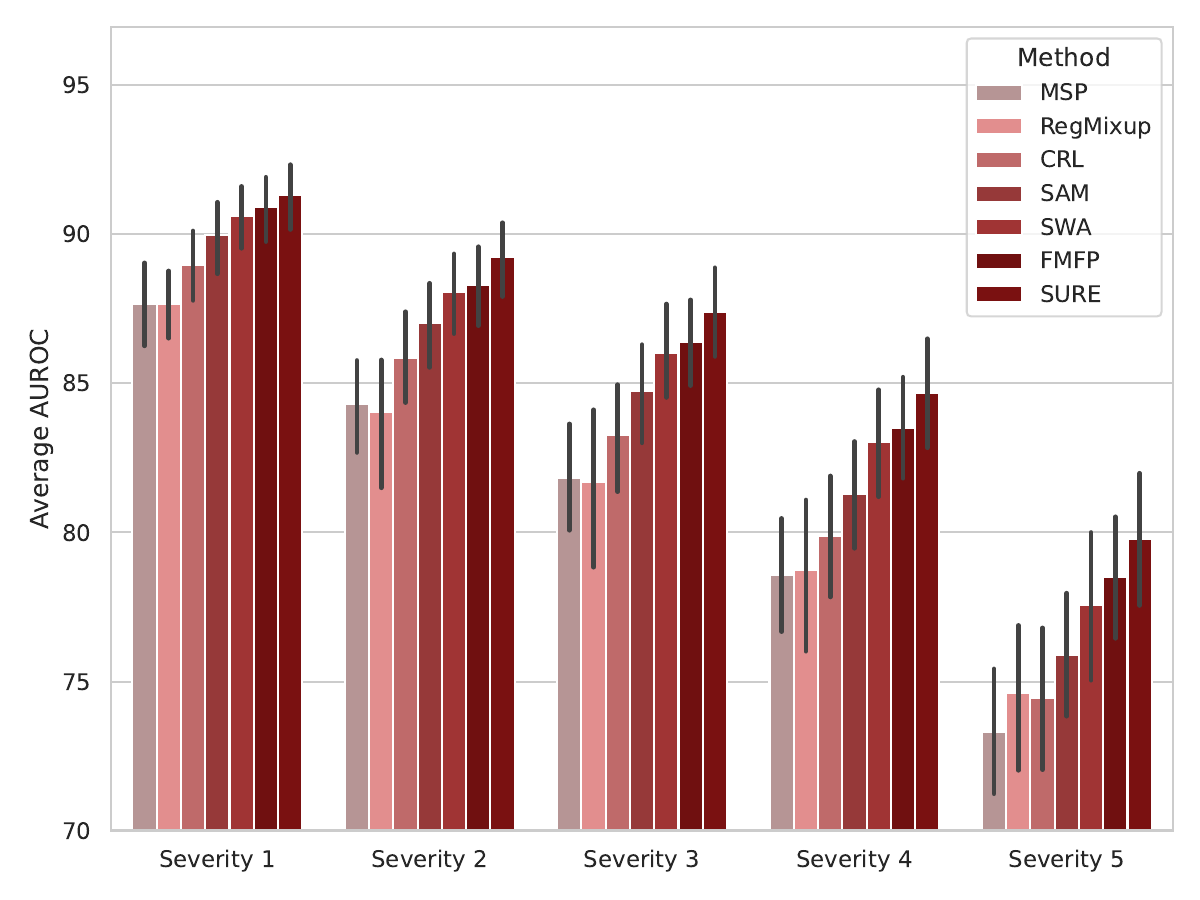}
    \caption{AUROC}
\end{subfigure}

\begin{subfigure}{0.46\textwidth}
    \centering
    \includegraphics[width=0.9\textwidth]{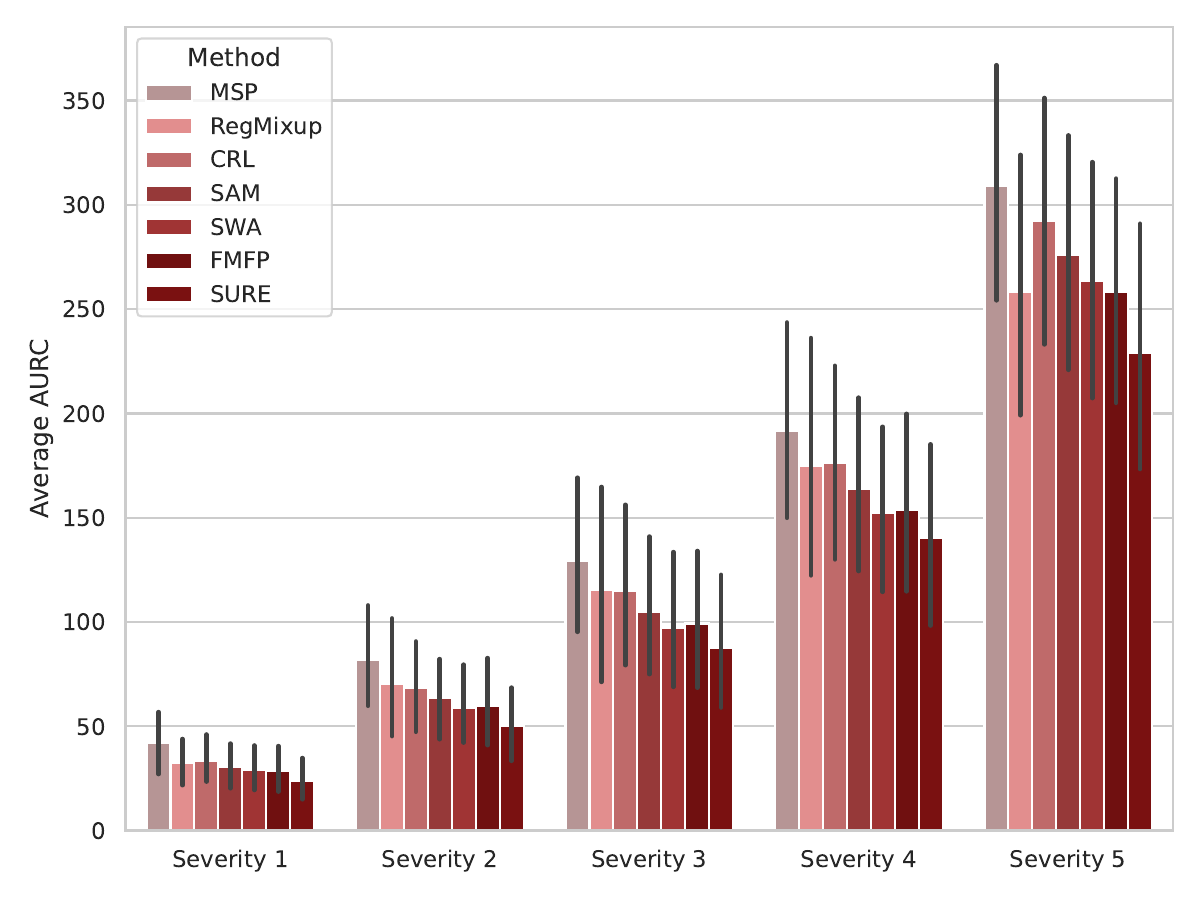}
    \caption{AURC}
\end{subfigure}
\vspace{-2mm}
\caption{\textbf{Comparison of the average AUROC \cite{davis2006relationship} (higher is better) and AURC \cite{geifman2018bias} (lower is better) on CIFAR10-C \cite{hendrycks2019benchmarking}.} We use DenseNet \cite{huang2017densely} as the backbone and train on the standard CIFAR10 training set. The evaluation results are averaged across the images with 15 types of corruption under 5 severity levels.}
\label{fig::combined cifarc}
\end{figure}

\paragraph{Comparison to state-of-the-art approaches} We also conduct fair comparison to state-of-the-art approaches on CIFAR10-LT \cite{cui2019class} and CIFAR100-LT \cite{cui2019class} with different imbalance factors. To make a fair comparison in this task, we train our SURE with ResNet32 \cite{he2016deep}, the most commonly used backbone in the community. The results are presented in Table~\ref{tab::longtail_results}. Although our proposed SURE is not originally designed for long-tailed classification, it achieves competitive results by equipping with the second stage uncertainty-aware re-weighting compared to task-specific solutions. The results suggest that leveraging uncertainty estimation for downstream applications is promising, especially using SURE to train the DNNs.

\subsection{Learning with noisy labels}
\label{sec::labelnoise}
For learning with noisy labels, we report top-1 test accuracy on benchmark Animal-10N~\cite{song2019selfie} and Food-101N~\cite{lee2018cleannet} in Tables~\ref{tab::animal10n} and~\ref{tab::food101n} , respectively. On Animal-10N, our SURE outperforms the baseline trained with cross-entropy loss by 9.6\%. Compared with NCT~\cite{chen2022compressing}, which uses two backbones for training, SURE trained with only one backbone improves performance by 4.9\%. Moreover, SURE achieves higher accuracy than SSR+~\cite{feng2022ssr}, which is designed for noisy labels employing techniques such as sample selection and relabelling.
In Table~\ref{tab::food101n} on Food-101N, although SURE is not designed for learning with noisy labels, with the default settings, it significantly outperforms all current SOTAs by at least 1.3\%. Results on both benchmarks verify SURE's robustness towards datasets with label noise.

\subsection{Failure prediction under distribution shift}
\label{sec::corruption}
In real-world applications, environmental conditions are prone to change frequently, such as shifts in weather from sunny to cloudy and then to rainy. It's crucial for models to maintain reliable decision-making capabilities under such distribution or domain shifts. To emulate these scenarios, we evaluate our model trained with the clean training set of CIFAR10 (the same training set presents in Section~\ref{sec::failure prediction}) on corruption datasets CIFAR10-C~\cite{hendrycks2019benchmarking}. We present the average AUROC and AURC of 15 corruptions for different approaches in Figure \ref{fig::combined cifarc}. Our SURE significantly enhances the failure prediction performance across a spectrum of corruptions. When compared to our baseline model, the SURE-based model demonstrates a notable improvement: the average AURC is reduced from 309 to 229. These results highlight SURE's robustness and adaptability in dynamically changing environments. Note that the performances of each corruption are presented in Appendix \ref{section::corruption}.

\begin{table}[!t]
    \begin{center}
		\resizebox{\columnwidth}{!}{
			\begin{tabular}{|c|c|c|c|c|c|c|c|c|c|c|}
				\hline \hline 
			Method	&\multicolumn{2}{|c|} {Loss}&\multicolumn{2}{|c|} {Optimzation}&\multicolumn{1}{|c|} {Classifer} &\multicolumn{4}{c|}{CIFAR100 \cite{krizhevsky2009learning}}  \\ 
                    \hline
                        &$\weightcrl$ &$\weightmixup$ &SAM&SWA& CSC&Acc.  $\uparrow$  &AURC $\downarrow$ &AUROC $\uparrow$ & FPR95 $\downarrow$                                
                        \\ 
			         
				   Baseline(MSP) &0&0&\XSolidBrush&\XSolidBrush&\XSolidBrush & 75.87±0.31 &69.44±2.11 & 87.00±0.21 & 60.73±1.16\\
				      SAM&0&0&\Checkmark&\XSolidBrush&\XSolidBrush  & 76.60±0.21	&62.97±1.02&	87.72±0.10&	59.35±0.87
                        \\
                         \textcolor{blue}{SWA}&0&0&\XSolidBrush&\Checkmark&\XSolidBrush & \textcolor{blue}{77.65±0.19}	&\textcolor{blue}{55.87±0.32}&	\textcolor{blue}{88.55±0.25}	&\textcolor{blue}{60.43±1.90}\\
                         CSC&0&0&\XSolidBrush&\XSolidBrush&\Checkmark &74.05±0.18	&78.14±0.26	&86.82±0.24	&63.56±1.20\\
                        FMFP&0&0&\Checkmark&\Checkmark&\XSolidBrush & 
                        77.82±0.08&55.03±0.52	&88.59±0.07	&59.79±0.31\\
                       SAM + CSC&0&0&\Checkmark&\XSolidBrush&\Checkmark &75.97±0.39&	64.20±1.55	&88.06±0.19&59.36±1.21\\	
				     SWA + CSC&0&0&\Checkmark&\Checkmark& \XSolidBrush
         & 78.46±0.33&55.68±0.41	&87.74±0.44	&61.22±2.54 \\
          FMFP + CSC &0&0&\Checkmark&\Checkmark&\Checkmark 
          & 78.45±0.13&54.18±0.47	&88.23±0.20	&60.05±1.03\\
          
           CRL &1&0&\XSolidBrush&\XSolidBrush&\XSolidBrush & 76.42±0.21&	62.78±0.21&88.07±0.17&59.02±0.39\\
            CRL + SAM &1&0&\Checkmark&\XSolidBrush&\XSolidBrush  & 76.98±0.32&	59.71±1.39&	88.26±0.07&	59.52±1.92\\
             CRL + SWA &1&0&\XSolidBrush&\Checkmark&\XSolidBrush & 
             77.56±0.20&56.88±0.28&88.24±0.45&61.73±1.77\\
              CRL + CSC &1&0&\XSolidBrush&\XSolidBrush&\Checkmark & 75.61±0.46&67.83±1.98&87.84±0.11&59.80±2.16\\
            CRL + FMFP &1&0&\Checkmark&\Checkmark&\XSolidBrush & 77.71±0.54&	56.24±0.89	&88.21±0.44	&61.75±1.74\\
               CRL+ SAM + CSC &1&0&\Checkmark&\XSolidBrush&\Checkmark & 78.21±0.53	&53.55±3.28&	88.86±0.45	&56.37±1.71\\
                 CRL+ SWA + CSC &1&0&\Checkmark&\Checkmark& \XSolidBrush & 78.09±0.10&56.61±0.91&	87.78±0.21&	61.37±1.56\\
                  CRL+ FMFP + CSC &1&0&\Checkmark&\Checkmark&\Checkmark & 78.24±0.18&	55.01±0.44&	88.14±0.11&	60.48±0.27\\
                  
            \textcolor{blue}{Reg} &0&1&\XSolidBrush&\XSolidBrush&\XSolidBrush & \textcolor{blue}{76.99±1.19}	& \textcolor{blue}{63.09±4.22}	&\textcolor{blue}{87.71±0.13}	&\textcolor{blue}{58.78±0.50}\\
            Reg + SAM &0&1&\Checkmark&\XSolidBrush&\XSolidBrush  & 77.45±0.55	&60.68±3.75&	87.70±0.39&	58.72±1.42\\
            Reg + SWA &0&1&\XSolidBrush&\Checkmark&\XSolidBrush & 78.55±0.62	&52.31±2.10&	88.71±0.22&	58.99±2.07\\
            \textcolor{blue}{Reg + CSC} &0&1&\XSolidBrush&\XSolidBrush&\Checkmark & \textcolor{blue}{78.32±0.28}	& \textcolor{blue}{62.40±0.58}& \textcolor{blue}{86.57±0.34}&\textcolor{blue}{58.77±2.27}\\
            \textcolor{blue}{Reg + FMFP} &0&1&\Checkmark&\Checkmark&\XSolidBrush & \textcolor{blue}{79.04±0.50}	& \textcolor{blue}{50.09±1.00}&	\textcolor{blue}{88.89±0.20}&	\textcolor{blue}{58.47±0.88}\\
            Reg + SAM + CSC &0&1&\Checkmark&\XSolidBrush&\Checkmark & 78.91±0.34&57.43±2.25&87.16±0.23&58.35±0.22\\
            Reg + SWA + CSC &0&1&\Checkmark&\Checkmark& \XSolidBrush & 80.17±0.52&49.87±1.86&87.89±0.10&61.08±1.06\\
            Reg + FMFP + CSC &0&1&\Checkmark&\Checkmark&\Checkmark &79.88±0.07	&48.58±0.34	&88.50±0.20	&58.52±0.75\\

            \textcolor{blue}{CRL + Reg} &1&1&\XSolidBrush&\XSolidBrush&\XSolidBrush & \textcolor{blue}{78.38±0.17}	& \textcolor{blue}{52.93±1.19}	& \textcolor{blue}{\bf88.97±0.38}	& \textcolor{blue}{\bf56.12±1.33} \\
            CRL + Reg + SAM &1&1&\Checkmark&\XSolidBrush&\XSolidBrush  & 78.21±0.53	&53.55±3.28	&88.86±0.45	&56.37±1.71\\
            CRL + Reg + SWA &1&1&\XSolidBrush&\Checkmark&\XSolidBrush & 78.64±0.16	&50.96±1.01&	88.96±0.31	&59.27±1.47\\
            CRL + Reg + CSC &1&1&\XSolidBrush&\XSolidBrush&\Checkmark & 79.42±0.11	&54.35±0.91&	87.59±0.20	&59.67±0.53\\
            CRL + Reg + FMFP &1&1&\Checkmark&\Checkmark&\XSolidBrush & 79.17±0.30	&49.96±1.63	&88.70±0.20&	59.85±2.07\\
            CRL + Reg + SAM + CSC &1&1&\Checkmark&\XSolidBrush&\Checkmark & 79.10±0.34	&56.39±1.25	&87.44±0.16	&56.98±0.31\\
            CRL + Reg + SWA + CSC &1&1&\Checkmark&\Checkmark& \XSolidBrush & 79.63±0.27 &49.14±0.22	&88.51±0.34	&59.28±2.14\\
            \textcolor{red}{\bf SURE} &1&1&\Checkmark&\Checkmark&\Checkmark & 
            \textcolor{red}{\bf 80.49±0.18}& \textcolor{red}{\bf 45.81±0.15}&\textcolor{red}{88.73±0.24}&\textcolor{red}{58.91±0.58}\\
                     \hline \hline
 \end{tabular}}
            
		\caption{\textbf{Ablation study} of \textbf{different components} used in SURE and their \textbf{combinations} on CIFAR100 \cite{krizhevsky2009learning}.}
  \label{tab::more ablations}
	\end{center}
\end{table}

\subsection{Analysis}
\label{section::ablation}
\paragraph{Ablation study} To further analyze SURE, we analyze the contribution of each component to our model's performance on CIFAR100 in Table \ref{tab::more ablations}. We report the means and standard deviations over \textit{three} runs in our ablations with ResNet18 \cite{he2016deep}. Starting from our baseline model, MSP, we observe the incremental impact of adding techniques like RegMixup, CRL, SAM, SWA, and the CSC to the SURE framework. Each addition to the SURE approach appears to improve accuracy and AURC, with the complete SURE method achieving the highest scores reported in the study. Among them, RegMixup and SWA contribute the most to performance, 
the combination of RegMixup and FMFP holds importance. This comprehensive analysis highlights the synergistic effect of our model's components, underscoring their collective importance in achieving optimal performance. Note that more analysis, such as the effect of RegMixup regularization weight $\weightmixup$ and CRL weight $\weightcrl$ are provided in Appendix \ref{section::lambda}.

\begin{figure}
\centering
\includegraphics[width=\linewidth]{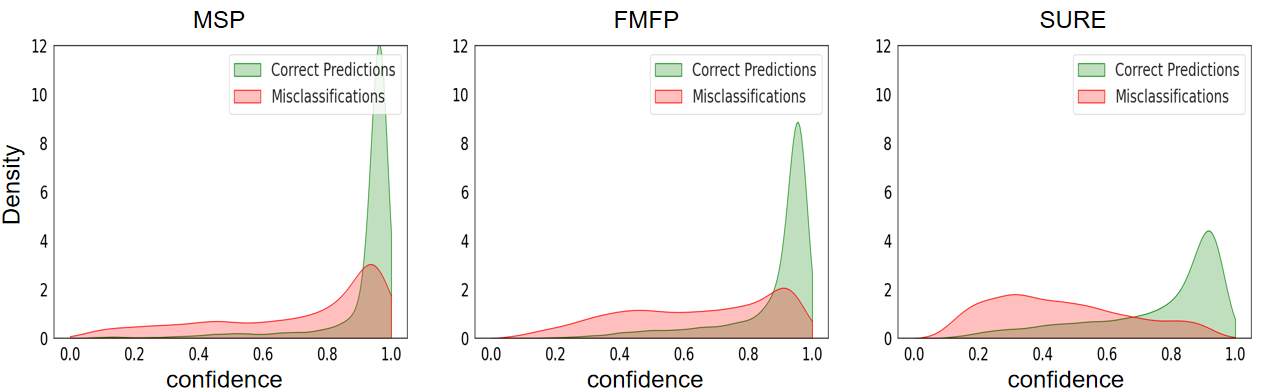}
\caption{\textbf{The visual results of confidence separation given by different methods on CIFAR100-LT \cite{cui2019class} IF=10.} SURE leads to better confidence separation than MSP \cite{hendrycks2016baseline} and FMFP \cite{zhu2022rethinking}.} 
\label{figure::confidence separation}
\end{figure}

\paragraph{Visualization}
\label{section::visual results}
We provide visualization of confidence distribution on CIFAR100-LT \cite{cui2019class} IF=10 in Figure \ref{figure::confidence separation}. From which, one can recognize SURE leads to clearly better confidence separation than MSP and FMFP. From the competitive approaches to SURE, the proposed method can increase the uncertainty of misclassified samples while improving accuracy.

\section{Conclusion}
In this paper, we introduce SURE, a novel framework that integrates multiple techniques for model regularization, classifier and optimization, aiming to enhance the reliability and robustness of DNNs. Our work highlights the shortcomings of existing methods when dealing with the complex nature of real-world data. This insight underlines the imperative need for approaches like SURE. Through rigorous evaluation, SURE has consistently outperformed individual methods across various datasets and model architectures in failure prediction. Moreover, its application in addressing real-world challenges, such as long-tailed classification, learning with noisy labels and data corruption, has not only yielded
results comparable to state-of-the-art methods in long-tailed distribution datasets but also excelled in scenarios with label noise. This work paves the way for the application of uncertainty estimation methods in various intricate real-world situations.  

\paragraph{Acknowledgement} We thank Caizhi Zhu, Yuming Du and Yinqiang Zheng for inspiring discussions and valuable feedback.

\clearpage
\newpage
{   
    \small
    \bibliographystyle{ieeenat_fullname}
    \bibliography{main}

\begin{thebibliography}{83}
\providecommand{\natexlab}[1]{#1}
\providecommand{\url}[1]{\texttt{#1}}
\expandafter\ifx\csname urlstyle\endcsname\relax
  \providecommand{\doi}[1]{doi: #1}\else
  \providecommand{\doi}{doi: \begingroup \urlstyle{rm}\Url}\fi

\bibitem[Alshammari et~al.(2022)Alshammari, Wang, Ramanan, and Kong]{alshammari2022long}
Shaden Alshammari, Yu-Xiong Wang, Deva Ramanan, and Shu Kong.
\newblock Long-tailed recognition via weight balancing.
\newblock In \emph{CVPR}, 2022.

\bibitem[Ayhan et~al.(2020)Ayhan, K{\"u}hlewein, Aliyeva, Inhoffen, Ziemssen, and Berens]{ayhan2020expert}
Murat~Se{\c{c}}kin Ayhan, Laura K{\"u}hlewein, Gulnar Aliyeva, Werner Inhoffen, Focke Ziemssen, and Philipp Berens.
\newblock Expert-validated estimation of diagnostic uncertainty for deep neural networks in diabetic retinopathy detection.
\newblock \emph{Medical image analysis}, 2020.

\bibitem[Berthelot et~al.(2019)Berthelot, Carlini, Goodfellow, Papernot, Oliver, and Raffel]{berthelot2019mixmatch}
David Berthelot, Nicholas Carlini, Ian Goodfellow, Nicolas Papernot, Avital Oliver, and Colin~A Raffel.
\newblock Mixmatch: A holistic approach to semi-supervised learning.
\newblock \emph{NeurIPS}, 32, 2019.

\bibitem[Cao et~al.(2019)Cao, Wei, Gaidon, Arechiga, and Ma]{cao2019learning}
Kaidi Cao, Colin Wei, Adrien Gaidon, Nikos Arechiga, and Tengyu Ma.
\newblock Learning imbalanced datasets with label-distribution-aware margin loss.
\newblock In \emph{NeurIPS}, 2019.

\bibitem[Chang et~al.(2017)Chang, Learned-Miller, and McCallum]{chang2017active}
Haw-Shiuan Chang, Erik Learned-Miller, and Andrew McCallum.
\newblock Active bias: Training more accurate neural networks by emphasizing high variance samples.
\newblock In \emph{NeurIPS}, 2017.

\bibitem[Chen et~al.(2022)Chen, Hu, Shen, Ai, and Suykens]{chen2022compressing}
Yingyi Chen, Shell~Xu Hu, Xi Shen, Chunrong Ai, and Johan A.~K. Suykens.
\newblock Compressing features for learning with noisy labels.
\newblock \emph{TNNLS}, 2022.

\bibitem[Chen et~al.(2023{\natexlab{a}})Chen, Shen, Liu, Tao, and Suykens]{chen2023jigsaw}
Yingyi Chen, Xi Shen, Yahui Liu, Qinghua Tao, and Johan~AK Suykens.
\newblock Jigsaw-vit: Learning jigsaw puzzles in vision transformer.
\newblock \emph{Pattern Recognition Letters}, 2023{\natexlab{a}}.

\bibitem[Chen et~al.(2023{\natexlab{b}})Chen, Zhang, Kou, Chen, Hsieh, and Gu]{chen2023does}
Zixiang Chen, Junkai Zhang, Yiwen Kou, Xiangning Chen, Cho-Jui Hsieh, and Quanquan Gu.
\newblock Why does sharpness-aware minimization generalize better than sgd?
\newblock In \emph{NeurIPS}, 2023{\natexlab{b}}.

\bibitem[Choi et~al.(2019)Choi, Chun, Kim, and Lee]{choi2019gaussian}
Jiwoong Choi, Dayoung Chun, Hyun Kim, and Hyuk-Jae Lee.
\newblock Gaussian yolov3: An accurate and fast object detector using localization uncertainty for autonomous driving.
\newblock In \emph{ICCV}, 2019.

\bibitem[Corbiere et~al.(2021)Corbiere, Thome, Saporta, Vu, Cord, and Perez]{corbiere2021confidence}
Charles Corbiere, Nicolas Thome, Antoine Saporta, Tuan-Hung Vu, Matthieu Cord, and Patrick Perez.
\newblock Confidence estimation via auxiliary models.
\newblock \emph{IEEE TPAMI}, 2021.

\bibitem[Cui et~al.(2022)Cui, Liu, Tian, Zhong, and Jia]{cui2022reslt}
Jiequan Cui, Shu Liu, Zhuotao Tian, Zhisheng Zhong, and Jiaya Jia.
\newblock Reslt: Residual learning for long-tailed recognition.
\newblock \emph{IEEE TPAMI}, 2022.

\bibitem[Cui et~al.(2019)Cui, Jia, Lin, Song, and Belongie]{cui2019class}
Yin Cui, Menglin Jia, Tsung-Yi Lin, Yang Song, and Serge Belongie.
\newblock Class-balanced loss based on effective number of samples.
\newblock In \emph{CVPR}, 2019.

\bibitem[Davis and Goadrich(2006)]{davis2006relationship}
Jesse Davis and Mark Goadrich.
\newblock The relationship between precision-recall and roc curves.
\newblock In \emph{ICML}, 2006.

\bibitem[Deng et~al.(2009)Deng, Dong, Socher, Li, Li, and Fei-Fei]{deng2009imagenet}
Jia Deng, Wei Dong, Richard Socher, Li-Jia Li, Kai Li, and Li Fei-Fei.
\newblock Imagenet: A large-scale hierarchical image database.
\newblock In \emph{CVPR}, 2009.

\bibitem[Ding et~al.(2018)Ding, Wang, Fan, and Gong]{ding2018semi}
Yifan Ding, Liqiang Wang, Deliang Fan, and Boqing Gong.
\newblock A semi-supervised two-stage approach to learning from noisy labels.
\newblock In \emph{WACV}, 2018.

\bibitem[Du et~al.(2023)Du, Yang, Jia, Nan, Chen, and Yang]{du2023global}
Fei Du, Peng Yang, Qi Jia, Fengtao Nan, Xiaoting Chen, and Yun Yang.
\newblock Global and local mixture consistency cumulative learning for long-tailed visual recognitions.
\newblock In \emph{CVPR}, 2023.

\bibitem[Feng et~al.(2022)Feng, Tzimiropoulos, and Patras]{feng2022ssr}
Chen Feng, Georgios Tzimiropoulos, and Ioannis Patras.
\newblock Ssr: An efficient and robust framework for learning with unknown label noise.
\newblock In \emph{BMVC}, 2022.

\bibitem[Feng et~al.(2018)Feng, Rosenbaum, and Dietmayer]{feng2018towards}
Di Feng, Lars Rosenbaum, and Klaus Dietmayer.
\newblock Towards safe autonomous driving: Capture uncertainty in the deep neural network for lidar 3d vehicle detection.
\newblock In \emph{2018 21st international conference on intelligent transportation systems (ITSC)}, 2018.

\bibitem[Foret et~al.(2020)Foret, Kleiner, Mobahi, and Neyshabur]{foret2020sharpness}
Pierre Foret, Ariel Kleiner, Hossein Mobahi, and Behnam Neyshabur.
\newblock Sharpness-aware minimization for efficiently improving generalization.
\newblock In \emph{ICLR}, 2020.

\bibitem[Franchi et~al.(2022)Franchi, Yu, Bursuc, Aldea, Dubuisson, and Filliat]{franchi2022latent}
Gianni Franchi, Xuanlong Yu, Andrei Bursuc, Emanuel Aldea, Severine Dubuisson, and David Filliat.
\newblock Latent discriminant deterministic uncertainty.
\newblock In \emph{ECCV}, 2022.

\bibitem[Geifman and El-Yaniv(2017)]{geifman2017selective}
Yonatan Geifman and Ran El-Yaniv.
\newblock Selective classification for deep neural networks.
\newblock In \emph{NeurIPS}, 2017.

\bibitem[Geifman et~al.(2018)Geifman, Uziel, and El-Yaniv]{geifman2018bias}
Yonatan Geifman, Guy Uziel, and Ran El-Yaniv.
\newblock Bias-reduced uncertainty estimation for deep neural classifiers.
\newblock In \emph{ICLR}, 2018.

\bibitem[Gidaris and Komodakis(2018)]{gidaris2018dynamic}
Spyros Gidaris and Nikos Komodakis.
\newblock Dynamic few-shot visual learning without forgetting.
\newblock In \emph{CVPR}, 2018.

\bibitem[Giuffrida et~al.(2021)Giuffrida, Fanucci, Meoni, Bati{\v{c}}, Buckley, Dunne, van Dijk, Esposito, Hefele, Vercruyssen, et~al.]{giuffrida2021varphi}
Gianluca Giuffrida, Luca Fanucci, Gabriele Meoni, Matej Bati{\v{c}}, L{\'e}onie Buckley, Aubrey Dunne, Chris van Dijk, Marco Esposito, John Hefele, Nathan Vercruyssen, et~al.
\newblock The $\phi$-sat-1 mission: The first on-board deep neural network demonstrator for satellite earth observation.
\newblock \emph{IEEE Transactions on Geoscience and Remote Sensing}, 2021.

\bibitem[Guo et~al.(2017)Guo, Pleiss, Sun, and Weinberger]{guo2017calibration}
Chuan Guo, Geoff Pleiss, Yu Sun, and Kilian~Q Weinberger.
\newblock On calibration of modern neural networks.
\newblock In \emph{ICML}, 2017.

\bibitem[Han et~al.(2018)Han, Yao, Yu, Niu, Xu, Hu, Tsang, and Sugiyama]{han2018co}
Bo Han, Quanming Yao, Xingrui Yu, Gang Niu, Miao Xu, Weihua Hu, Ivor Tsang, and Masashi Sugiyama.
\newblock Co-teaching: Robust training of deep neural networks with extremely noisy labels.
\newblock In \emph{NeurIPS}, 2018.

\bibitem[Han et~al.(2019)Han, Luo, and Wang]{han2019deep}
Jiangfan Han, Ping Luo, and Xiaogang Wang.
\newblock Deep self-learning from noisy labels.
\newblock In \emph{ICCV}, 2019.

\bibitem[He et~al.(2016)He, Zhang, Ren, and Sun]{he2016deep}
Kaiming He, Xiangyu Zhang, Shaoqing Ren, and Jian Sun.
\newblock Deep residual learning for image recognition.
\newblock In \emph{CVPR}, 2016.

\bibitem[He et~al.(2015)He, Chen, and Yin]{he2015adaptive}
Wei He, Yuhao Chen, and Zhao Yin.
\newblock Adaptive neural network control of an uncertain robot with full-state constraints.
\newblock \emph{IEEE transactions on cybernetics}, 2015.

\bibitem[Hendrycks and Dietterich(2019)]{hendrycks2019benchmarking}
Dan Hendrycks and Thomas Dietterich.
\newblock Benchmarking neural network robustness to common corruptions and perturbations.
\newblock In \emph{ICLR}, 2019.

\bibitem[Hendrycks and Gimpel(2017)]{hendrycks2016baseline}
Dan Hendrycks and Kevin Gimpel.
\newblock A baseline for detecting misclassified and out-of-distribution examples in neural networks.
\newblock In \emph{ICLR}, 2017.

\bibitem[Hendrycks et~al.(2018)Hendrycks, Mazeika, and Dietterich]{hendrycks2018deep}
Dan Hendrycks, Mantas Mazeika, and Thomas Dietterich.
\newblock Deep anomaly detection with outlier exposure.
\newblock \emph{ICLR}, 2018.

\bibitem[Hu et~al.(2020)Hu, Moreno, Xiao, Shen, Obozinski, Lawrence, and Damianou]{hu2020empirical}
Shell~Xu Hu, Pablo~G Moreno, Yang Xiao, Xi Shen, Guillaume Obozinski, Neil~D Lawrence, and Andreas Damianou.
\newblock Empirical bayes transductive meta-learning with synthetic gradients.
\newblock In \emph{ICLR}, 2020.

\bibitem[Huang et~al.(2017)Huang, Liu, Van Der~Maaten, and Weinberger]{huang2017densely}
Gao Huang, Zhuang Liu, Laurens Van Der~Maaten, and Kilian~Q Weinberger.
\newblock Densely connected convolutional networks.
\newblock In \emph{CVPR}, 2017.

\bibitem[Izmailov et~al.(2018)Izmailov, Podoprikhin, Garipov, Vetrov, and Wilson]{izmailov2018averaging}
Pavel Izmailov, Dmitrii Podoprikhin, Timur Garipov, Dmitry Vetrov, and Andrew~Gordon Wilson.
\newblock Averaging weights leads to wider optima and better generalization.
\newblock \emph{arXiv}, 2018.

\bibitem[Jiang et~al.(2018)Jiang, Zhou, Leung, Li, and Fei-Fei]{jiang2018mentornet}
Lu Jiang, Zhengyuan Zhou, Thomas Leung, Li-Jia Li, and Li Fei-Fei.
\newblock {M}entor{N}et: Learning data-driven curriculum for very deep neural networks on corrupted labels.
\newblock In \emph{ICML}, 2018.

\bibitem[Jiang et~al.(2023)Jiang, Li, Zhang, Wang, and Xu]{jiang2023dynamic}
Shenwang Jiang, Jianan Li, Jizhou Zhang, Ying Wang, and Tingfa Xu.
\newblock Dynamic loss for robust learning.
\newblock \emph{IEEE TPAMI}, 2023.

\bibitem[Kang et~al.(2020)Kang, Xie, Rohrbach, Yan, Gordo, Feng, and Kalantidis]{kang2019decoupling}
Bingyi Kang, Saining Xie, Marcus Rohrbach, Zhicheng Yan, Albert Gordo, Jiashi Feng, and Yannis Kalantidis.
\newblock Decoupling representation and classifier for long-tailed recognition.
\newblock In \emph{ICLR}, 2020.

\bibitem[Kong et~al.(2019)Kong, Lee, Kwak, Kang, Kim, and Song]{kong2019recycling}
Kyeongbo Kong, Junggi Lee, Youngchul Kwak, Minsung Kang, Seong~Gyun Kim, and Woo-Jin Song.
\newblock Recycling: Semi-supervised learning with noisy labels in deep neural networks.
\newblock \emph{IEEE Access}, 7:\penalty0 66998--67005, 2019.

\bibitem[Krizhevsky et~al.(2009)Krizhevsky, Hinton, et~al.]{krizhevsky2009learning}
Alex Krizhevsky, Geoffrey Hinton, et~al.
\newblock Learning multiple layers of features from tiny images.
\newblock \emph{Toronto, ON, Canada}, 2009.

\bibitem[Le and Yang(2015)]{le2015tiny}
Ya Le and Xuan Yang.
\newblock Tiny imagenet visual recognition challenge.
\newblock \emph{CS 231N}, 7\penalty0 (7):\penalty0 3, 2015.

\bibitem[Lee et~al.(2018)Lee, He, Zhang, and Yang]{lee2018cleannet}
Kuang-Huei Lee, Xiaodong He, Lei Zhang, and Linjun Yang.
\newblock Cleannet: Transfer learning for scalable image classifier training with label noise.
\newblock In \emph{CVPR}, 2018.

\bibitem[Leibig et~al.(2017)Leibig, Allken, Ayhan, Berens, and Wahl]{leibig2017leveraging}
Christian Leibig, Vaneeda Allken, Murat~Se{\c{c}}kin Ayhan, Philipp Berens, and Siegfried Wahl.
\newblock Leveraging uncertainty information from deep neural networks for disease detection.
\newblock \emph{Scientific reports}, 2017.

\bibitem[Li et~al.(2020)Li, Socher, and Hoi]{li2020dividemix}
Junnan Li, Richard Socher, and Steven~CH Hoi.
\newblock Dividemix: Learning with noisy labels as semi-supervised learning.
\newblock In \emph{ICLR}, 2020.

\bibitem[Li et~al.(2021)Li, Gong, Liu, Wang, Qiao, and Cheng]{li2021metasaug}
Shuang Li, Kaixiong Gong, Chi~Harold Liu, Yulin Wang, Feng Qiao, and Xinjing Cheng.
\newblock Metasaug: Meta semantic augmentation for long-tailed visual recognition.
\newblock In \emph{CVPR}, 2021.

\bibitem[Liang et~al.(2017)Liang, Li, and Srikant]{liang2017enhancing}
Shiyu Liang, Yixuan Li, and Rayadurgam Srikant.
\newblock Enhancing the reliability of out-of-distribution image detection in neural networks.
\newblock \emph{ICLR}, 2017.

\bibitem[Liu et~al.(2020{\natexlab{a}})Liu, Niles-Weed, Razavian, and Fernandez-Granda]{liu2020early}
Sheng Liu, Jonathan Niles-Weed, Narges Razavian, and Carlos Fernandez-Granda.
\newblock Early-learning regularization prevents memorization of noisy labels.
\newblock In \emph{NeurIPS}, 2020{\natexlab{a}}.

\bibitem[Liu et~al.(2020{\natexlab{b}})Liu, Wang, Owens, and Li]{liu2020energy}
Weitang Liu, Xiaoyun Wang, John Owens, and Yixuan Li.
\newblock Energy-based out-of-distribution detection.
\newblock \emph{NeurIPS}, 2020{\natexlab{b}}.

\bibitem[Loquercio et~al.(2020)Loquercio, Segu, and Scaramuzza]{loquercio2020general}
Antonio Loquercio, Mattia Segu, and Davide Scaramuzza.
\newblock A general framework for uncertainty estimation in deep learning.
\newblock \emph{IEEE Robotics and Automation Letters}, 2020.

\bibitem[Ma et~al.(2018)Ma, Wang, Houle, Zhou, Erfani, Xia, Wijewickrema, and Bailey]{ma2018dimensionality}
Xingjun Ma, Yisen Wang, Michael~E Houle, Shuo Zhou, Sarah Erfani, Shutao Xia, Sudanthi Wijewickrema, and James Bailey.
\newblock Dimensionality-driven learning with noisy labels.
\newblock In \emph{ICML}, 2018.

\bibitem[Malach and Shalev-Shwartz(2017)]{malach2017decoupling}
Eran Malach and Shai Shalev-Shwartz.
\newblock Decoupling ``when to update" from ``how to update".
\newblock In \emph{NeurIPS}, 2017.

\bibitem[Miralles et~al.(2023)Miralles, Thangavel, Scannapieco, Jagadam, Baranwal, Faldu, Abhang, Bhatia, Bonnart, Bhatnagar, et~al.]{miralles2023critical}
Pablo Miralles, Kathiravan Thangavel, Antonio~Fulvio Scannapieco, Nitya Jagadam, Prerna Baranwal, Bhavin Faldu, Ruchita Abhang, Sahil Bhatia, Sebastien Bonnart, Ishita Bhatnagar, et~al.
\newblock A critical review on the state-of-the-art and future prospects of machine learning for earth observation operations.
\newblock \emph{Advances in Space Research}, 2023.

\bibitem[Miyato et~al.(2018)Miyato, Kataoka, Koyama, and Yoshida]{miyato2018spectral}
Takeru Miyato, Toshiki Kataoka, Masanori Koyama, and Yuichi Yoshida.
\newblock Spectral normalization for generative adversarial networks.
\newblock \emph{ICLR}, 2018.

\bibitem[Moon et~al.(2020)Moon, Kim, Shin, and Hwang]{moon2020confidence}
Jooyoung Moon, Jihyo Kim, Younghak Shin, and Sangheum Hwang.
\newblock Confidence-aware learning for deep neural networks.
\newblock In \emph{ICML}, 2020.

\bibitem[Mukhoti et~al.(2023)Mukhoti, Kirsch, van Amersfoort, Torr, and Gal]{Mukhoti_2023_CVPR}
Jishnu Mukhoti, Andreas Kirsch, Joost van Amersfoort, Philip~H.S. Torr, and Yarin Gal.
\newblock Deep deterministic uncertainty: A new simple baseline.
\newblock In \emph{CVPR}, 2023.

\bibitem[Nair et~al.(2020)Nair, Precup, Arnold, and Arbel]{nair2020exploring}
Tanya Nair, Doina Precup, Douglas~L Arnold, and Tal Arbel.
\newblock Exploring uncertainty measures in deep networks for multiple sclerosis lesion detection and segmentation.
\newblock \emph{Medical image analysis}, 2020.

\bibitem[Nishi et~al.(2021)Nishi, Ding, Rich, and Hollerer]{nishi2021augmentation}
Kento Nishi, Yi Ding, Alex Rich, and Tobias Hollerer.
\newblock Augmentation strategies for learning with noisy labels.
\newblock In \emph{CVPR}, 2021.

\bibitem[Padhy et~al.(2020)Padhy, Nado, Ren, Liu, Snoek, and Lakshminarayanan]{padhy2020revisiting}
Shreyas Padhy, Zachary Nado, Jie Ren, Jeremiah Liu, Jasper Snoek, and Balaji Lakshminarayanan.
\newblock Revisiting one-vs-all classifiers for predictive uncertainty and out-of-distribution detection in neural networks.
\newblock In \emph{ICML Workshops}, 2020.

\bibitem[Pinto et~al.(2022)Pinto, Yang, Lim, Torr, and Dokania]{pinto2022using}
Francesco Pinto, Harry Yang, Ser~Nam Lim, Philip Torr, and Puneet Dokania.
\newblock Using mixup as a regularizer can surprisingly improve accuracy \& out-of-distribution robustness.
\newblock In \emph{NeurIPS}, 2022.

\bibitem[Pleiss et~al.(2020)Pleiss, Zhang, Elenberg, and Weinberger]{pleiss2020identifying}
Geoff Pleiss, Tianyi Zhang, Ethan Elenberg, and Kilian~Q Weinberger.
\newblock Identifying mislabeled data using the area under the margin ranking.
\newblock In \emph{NeurIPS}, 2020.

\bibitem[Qu et~al.(2023)Qu, Foo, Li, and Liu]{qu2023towards}
Haoxuan Qu, Lin~Geng Foo, Yanchao Li, and Jun Liu.
\newblock Towards more reliable confidence estimation.
\newblock \emph{IEEE TPAMI}, 2023.

\bibitem[Sharma et~al.(2020)Sharma, Donmez, Luo, Liu, and Yalniz]{sharma2020noiserank}
Karishma Sharma, Pinar Donmez, Enming Luo, Yan Liu, and I~Zeki Yalniz.
\newblock Noiserank: Unsupervised label noise reduction with dependence models.
\newblock In \emph{ECCV}, 2020.

\bibitem[Shu et~al.(2019)Shu, Xie, Yi, Zhao, Zhou, Xu, and Meng]{shu2019meta}
Jun Shu, Qi Xie, Lixuan Yi, Qian Zhao, Sanping Zhou, Zongben Xu, and Deyu Meng.
\newblock Meta-weight-net: Learning an explicit mapping for sample weighting.
\newblock In \emph{NeurIPS}, 2019.

\bibitem[Simonyan and Zisserman(2014)]{simonyan2014very}
Karen Simonyan and Andrew Zisserman.
\newblock Very deep convolutional networks for large-scale image recognition.
\newblock In \emph{CVPR}, 2014.

\bibitem[Song et~al.(2019)Song, Kim, and Lee]{song2019selfie}
Hwanjun Song, Minseok Kim, and Jae-Gil Lee.
\newblock Selfie: Refurbishing unclean samples for robust deep learning.
\newblock In \emph{ICML}, 2019.

\bibitem[Steinhardt and Liang(2016)]{steinhardt2016unsupervised}
Jacob Steinhardt and Percy~S Liang.
\newblock Unsupervised risk estimation using only conditional independence structure.
\newblock \emph{NeurIPS}, 2016.

\bibitem[Sun et~al.(2022)Sun, Guo, Wei, Han, and Yin]{sun2022learning}
Haoliang Sun, Chenhui Guo, Qi Wei, Zhongyi Han, and Yilong Yin.
\newblock Learning to rectify for robust learning with noisy labels.
\newblock \emph{Pattern Recognition}, 2022.

\bibitem[Tanaka et~al.(2018)Tanaka, Ikami, Yamasaki, and Aizawa]{tanaka2018joint}
Daiki Tanaka, Daiki Ikami, Toshihiko Yamasaki, and Kiyoharu Aizawa.
\newblock Joint optimization framework for learning with noisy labels.
\newblock In \emph{CVPR}, 2018.

\bibitem[Tang et~al.(2020)Tang, Huang, and Zhang]{tang2020long}
Kaihua Tang, Jianqiang Huang, and Hanwang Zhang.
\newblock Long-tailed classification by keeping the good and removing the bad momentum causal effect.
\newblock In \emph{NeurIPS}, 2020.

\bibitem[Touvron et~al.(2021)Touvron, Cord, Douze, Massa, Sablayrolles, and J{\'e}gou]{touvron2021training}
Hugo Touvron, Matthieu Cord, Matthijs Douze, Francisco Massa, Alexandre Sablayrolles, and Herv{\'e} J{\'e}gou.
\newblock Training data-efficient image transformers \& distillation through attention.
\newblock In \emph{ICML}, 2021.

\bibitem[Wang et~al.(2021)Wang, Han, Wei, Zhang, and Wang]{wang2021contrastive}
Peng Wang, Kai Han, Xiu-Shen Wei, Lei Zhang, and Lei Wang.
\newblock Contrastive learning based hybrid networks for long-tailed image classification.
\newblock In \emph{CVPR}, 2021.

\bibitem[Wei et~al.(2020)Wei, Feng, Chen, and An]{wei2020combating}
Hongxin Wei, Lei Feng, Xiangyu Chen, and Bo An.
\newblock Combating noisy labels by agreement: A joint training method with co-regularization.
\newblock In \emph{CVPR}, 2020.

\bibitem[Yang and Xu(2020)]{yang2020rethinking}
Yuzhe Yang and Zhi Xu.
\newblock Rethinking the value of labels for improving class-imbalanced learning.
\newblock In \emph{NeurIPS}, 2020.

\bibitem[Yi and Wu(2019)]{yi2019probabilistic}
Kun Yi and Jianxin Wu.
\newblock Probabilistic end-to-end noise correction for learning with noisy labels.
\newblock In \emph{CVPR}, 2019.

\bibitem[Yu et~al.(2019)Yu, Han, Yao, Niu, Tsang, and Sugiyama]{yu2019does}
Xingrui Yu, Bo Han, Jiangchao Yao, Gang Niu, Ivor Tsang, and Masashi Sugiyama.
\newblock How does disagreement help generalization against label corruption?
\newblock In \emph{ICML}, 2019.

\bibitem[Zagoruyko and Komodakis(2016)]{zagoruyko2016wide}
Sergey Zagoruyko and Nikos Komodakis.
\newblock Wide residual networks.
\newblock In \emph{BMVC}, 2016.

\bibitem[Zhang et~al.(2018)Zhang, Cisse, Dauphin, and Lopez-Paz]{zhang2017mixup}
Hongyi Zhang, Moustapha Cisse, Yann~N Dauphin, and David Lopez-Paz.
\newblock mixup: Beyond empirical risk minimization.
\newblock In \emph{ICLR}, 2018.

\bibitem[Zhang et~al.(2021)Zhang, Zheng, Wu, Goswami, and Chen]{zhang2021learning}
Yikai Zhang, Songzhu Zheng, Pengxiang Wu, Mayank Goswami, and Chao Chen.
\newblock Learning with feature-dependent label noise: A progressive approach.
\newblock In \emph{ICLR}, 2021.

\bibitem[Zheltonozhskii et~al.(2022)Zheltonozhskii, Baskin, Mendelson, Bronstein, and Litany]{zheltonozhskii2022contrast}
Evgenii Zheltonozhskii, Chaim Baskin, Avi Mendelson, Alex~M Bronstein, and Or Litany.
\newblock Contrast to divide: Self-supervised pre-training for learning with noisy labels.
\newblock In \emph{WACV}, 2022.

\bibitem[Zhou et~al.(2020)Zhou, Cui, Wei, and Chen]{zhou2020bbn}
Boyan Zhou, Quan Cui, Xiu-Shen Wei, and Zhao-Min Chen.
\newblock Bbn: Bilateral-branch network with cumulative learning for long-tailed visual recognition.
\newblock In \emph{CVPR}, 2020.

\bibitem[Zhu et~al.(2022{\natexlab{a}})Zhu, Cheng, Zhang, and Liu]{zhu2022rethinking}
Fei Zhu, Zhen Cheng, Xu-Yao Zhang, and Cheng-Lin Liu.
\newblock Rethinking confidence calibration for failure prediction.
\newblock In \emph{ECCV}, 2022{\natexlab{a}}.

\bibitem[Zhu et~al.(2023)Zhu, Cheng, Zhang, and Liu]{zhu2023openmix}
Fei Zhu, Zhen Cheng, Xu-Yao Zhang, and Cheng-Lin Liu.
\newblock Openmix: Exploring outlier samples for misclassification detection.
\newblock In \emph{CVPR}, 2023.

\bibitem[Zhu et~al.(2022{\natexlab{b}})Zhu, Wang, Chen, Chen, and Jiang]{zhu2022balanced}
Jianggang Zhu, Zheng Wang, Jingjing Chen, Yi-Ping~Phoebe Chen, and Yu-Gang Jiang.
\newblock Balanced contrastive learning for long-tailed visual recognition.
\newblock In \emph{CVPR}, 2022{\natexlab{b}}.

\end{thebibliography}
}

\clearpage
\newpage

\appendix
\vspace*{1em}{\centering\Large\bf%
Appendix
\vspace*{1.5em}}

This appendix contains the following sections: 
\begin{itemize}
        
        \item Section~\ref{section::lambda}: Ablation study of $\lambda_\text{mix}$ and $\lambda_\text{crl}$ for the RegMixup \cite{pinto2022using} loss  and Correctness Ranking Loss (CRL) \cite{moon2020confidence}.  \\
        \item Section~\ref{section::tau}: Ablation study of $\tau$ for the Cosine Similarity Classifier (CSC)~\cite{gidaris2018dynamic,hu2020empirical}.
        \\
        \item Section~\ref{section::long-tail}: Comparison of the performance of different uncertainty estimation methods on CIFAR10-LT and CIFAR100-LT \cite{krizhevsky2009learning} with imbalance factor 10. \\
	\item Section~\ref{section::auroc}: More details about the definition of Area Under the Receiver Operating Characteristic Curve (AUROC) \cite{davis2006relationship} and False Positive Rate at 95\% True Positive Rate (FPR95) as mentioned in Section \textbf{4.1} (c.f. line 388) in our paper.
        \\
        
        \item Section~\ref{section::corruption}: More results of failure prediction under distribution shift.
        \\
        \item Section~\ref{section::re-weighting}: Ablation study of different re-weighting maps.
        \\
\end{itemize}

\section{Impact of different $\lambda_\text{crl}$ and $\lambda_\text{mix}$ in RegMixup \cite{pinto2022using} loss  and Correctness Ranking Loss (CRL) \cite{moon2020confidence}}
\label{section::lambda}
In this section, we present the results of varying the parameters $\lambda_\text{crl}$ and $\lambda_\text{mix}$ in the loss function of \textbf{SURE}. The experimental results, obtained using a ResNet18 \cite{he2016deep} backbone and summarized in Table~\ref{tab::lambda_ablation}, indicate that different datasets require different optimal weights. Notably, all experiments across various backbones consistently utilized the same values of $\lambda_\text{crl}$ and $\lambda_\text{mix}$ in our paper. We determined the optimal settings as 0.5 for both $\lambda_\text{crl}$ and $\lambda_\text{mix}$ on CIFAR10 \cite{krizhevsky2009learning}, 1 for CIFAR100 \cite{krizhevsky2009learning}, and 2 for Tiny-ImageNet \cite{le2015tiny}. Specifically, when we fine-tuned DeiT \cite{touvron2021training}, we set $\lambda_\text{crl}$ to 0 and $\lambda_\text{mix}$ to 0.2 across three datasets. Particularly in our downstream task, we set $\lambda_\text{crl}$ to 0 and $\lambda_\text{mix}$ to 1 when addressing the challenges of long-tailed distribution data. And we set $\lambda_\text{crl}$ to 0.2 and $\lambda_\text{mix}$ to 1 when learning with noisy labels.
\begin{table}
    \begin{center}
		\resizebox{\columnwidth}{!}{
			\begin{tabular}{|l|c|c|c|c|c|c|c|}
				\hline \hline 
				\multicolumn{1}{|c|} {Ratios} &\multicolumn{2}{c|}{CIFAR10 \cite{krizhevsky2009learning}} &\multicolumn{2}{c|}{CIFAR100 \cite{krizhevsky2009learning} } &\multicolumn{2}{c|}{Tiny-ImageNet \cite{le2015tiny}} \\ 
                    \hline
                        &\multicolumn{1}{c} {Acc. $\uparrow$}   &\multicolumn{1}{c|} {AURC $\downarrow$}
                        &\multicolumn{1}{c}  {Acc. $\uparrow$} 
                        &\multicolumn{1}{c|} {AURC $\downarrow$}
                        &\multicolumn{1}{c}  {Acc. $\uparrow$} 
                        &\multicolumn{1}{c|} {AURC $\downarrow$}
                        \\ 
			        Baseline(MSP) &95.41 ± 0.15  & 4.89 ± 0.96  & 74.91 ± 0.25  & 74.87 ± 0.24 & 63.27±0.04 & 134.87±1.14\\
                    \hline
                    \multicolumn{7}{|c|}{CRL weight $\lambda_{\text{crl}}$} \\
                    \hline
				    0.1& 95.47±0.19& 4.60±0.26 &75.47±0.46 &  75.02±2.99 & 63.32±0.23 & 135.62±2.56 \\
				    0.2& 95.33±0.26& 4.13±0.64 &76.04±0.78 &  73.03±2.04 & 63.44±0.16 & 131.62±1.37 \\
                    0.5& \bf 95.33±0.14& \bf 3.98±0.20 &75.49±0.39 &  71.84±1.49 & 64.86±0.02 & 124.63±0.49 \\
                    1&  95.13±0.16 & 4.67±0.40 & \bf 76.10±0.43 & \bf69.05±2.48 & 65.29±0.14 & 117.33±1.08 \\
                    2&  93.99±0.08& 6.71±0.28  &75.30±0.36 &  72.40±1.48  & \bf 65.59±0.18 & \bf 116.61±0.47 \\
                    5&  91.58±0.18 & 13.29±0.33&71.98±0.55 &  91.42±2.15  & 62.66±0.17 & 136.03±0.94 \\
                    \hline
                    \multicolumn{7}{|c|}{RegMixup regularization weight $\lambda_{\text{mix}}$}  \\
                    \hline
                    0.1& 95.76±0.08 & 5.81±0.98 & 77.59±0.67 & 66.49±2.09  &  65.42±0.40 & 123.37±1.00  \\
				    0.2& 95.85±0.11 & 4.74±0.41 & 77.35±0.39 & 66.59±0.77 & 65.59±0.20 & 122.26±0.67 \\
                    0.5& \bf 96.23±0.10 & \bf 4.68±0.47 & 77.21±0.52 & 66.32±1.96 & 66.26±0.21 & 116.50±2.31 \\
                    1&   95.96±0.29&  7.04±0.92 & \bf 77.64±0.85 & \bf 63.88±5.22 & 66.00±0.22& 117.79±1.49\\
                    2&   96.03±0.07&  7.03±0.45 & 77.13±0.31 & 66.56±0.43 & \bf 66.26±0.12 &  \bf 113.40±1.31 \\
                    5&   95.83±0.23 & 6.17±1.74 & 77.52±0.95 & 63.40±6.22  & 65.40±2.06 & 119.34±12.49 \\
                        \hline \hline
            
		    \end{tabular}}
		\caption{\textbf{Ablation Study of hyper-parameters $\lambda_{\text{crl}}$ and $\lambda_{\text{mix}}$ in the loss function of SURE.} Experiments are implemented on CIFAR10, CIFAR100 \cite{krizhevsky2009learning} and Tiny-ImageNet \cite{le2015tiny} datasets.} 
		\label{tab::lambda_ablation}
	\end{center}
\end{table}
\section{Impact of different $\tau$ in Cosine Similarity Classifier (CSC)~\cite{gidaris2018dynamic,hu2020empirical}  }
\label{section::tau}
In the same vein as the previous ablation study for $\lambda_\text{crl}$ and $\lambda_\text{mix}$, we also conducted an analysis of the cosine similarity classifier temperature $\tau$ within the \textbf{SURE} framework. This study is detailed in Table~\ref{tab::tau_ablation}. For CIFAR10 \cite{krizhevsky2009learning} and CIFAR100 \cite{krizhevsky2009learning}, the best-performing temperature value was found to be $\tau=8$, while for Tiny-ImageNet \cite{le2015tiny}, a higher temperature of $\tau=16$ yielded superior results. Specifically, when we fine-tuned DeiT \cite{touvron2021training}, we set the temperature of $\tau=16$ on three datasets. Note that across all our downstream tasks, we consistently applied a temperature of $\tau=8$ . 

\begin{table}
    \begin{center}
		\resizebox{\columnwidth}{!}{
			\begin{tabular}{|l|c|c|c|c|c|c|c|}
				\hline \hline 
				\multicolumn{1}{|c|} {Ratios} &\multicolumn{2}{c|}{CIFAR10 \cite{krizhevsky2009learning}} &\multicolumn{2}{c|}{CIFAR100 \cite{krizhevsky2009learning} } & \multicolumn{2}{c|}{Tiny-ImageNet \cite{le2015tiny}} \\ 
                    \hline
                        &\multicolumn{1}{c} {Acc. $\uparrow$}   &\multicolumn{1}{c|} {AURC $\downarrow$}
                        &\multicolumn{1}{c}  {Acc. $\uparrow$} 
                        &\multicolumn{1}{c|} {AURC $\downarrow$}
                        &\multicolumn{1}{c}  {Acc. $\uparrow$} 
                        &\multicolumn{1}{c|} {AURC $\downarrow$}\\ 
			        Baseline(MSP) &95.41±0.15  & 4.89±0.96  & 74.91±0.25  & 74.87±0.24 & 63.27±0.04 & 134.87±1.14 \\
                    \hline
                    \multicolumn{7}{|c|}{cosine similarity classifier temperature $\tau$} \\
                    \hline
				   4 &96.29±0.01 & 2.44±0.04  &79.73±0.22& 53.71±0.16 & 64.86±0.14&  128.28±1.76\\
				   8 & \bf 96.65±0.07& \bf 2.13±0.03 & \bf 80.37±0.07& \bf 48.20±0.73 & 68.26±0.05& 99.76±0.59 \\
                    16  &  96.17±0.10 & 2.52±0.07 &79.90±0.35& 50.28±1.29 & \bf 69.03±0.05  & \bf 94.63±0.74  \\
                    32  & 96.20±0.10& 2.51±0.06 & 79.07±0.32 & 53.14±1.82 & 67.44±0.29 & 103.51±1.89\\                  
                        \hline \hline
            
		    \end{tabular}}
		\caption{\textbf{Ablation Study of hyper-parameters $\tau$ in Cosine Similarity Classifier (CSC) of SURE.} Experiments are implemented on CIFAR10, CIFAR100~\cite{krizhevsky2009learning} and Tiny-ImageNet \cite{le2015tiny} datasets.} 
		\label{tab::tau_ablation}
	\end{center}
\end{table}
\section{More results of failure prediction on CIFAR10-LT and CIFAR100-LT ~\cite{krizhevsky2009learning}}
\label{section::long-tail}
We evaluate the performance of failure prediction under imbalanced data distribution. The Acc.~and AURC are provided in Table~\ref{tab::longtail_failure_prediction} for imbalance factor IF = 10. We find that even under imbalanced data distribution, our SURE still significantly outperforms other approaches of failure prediction across different datasets and backbones, demonstrating its robustness under more challenging settings.
\section{Definition of AUROC \cite{davis2006relationship} and FPR95}
\label{section::auroc}
\paragraph{AUROC}
The area under the receiver operating characteristic curve (AUROC) measures the area under the curve drawn
by plotting the true positive(TP) rate against the false positive(FP)
rate.
\paragraph{FPR95}
FPR95 is the abbreviation of FPR-at-95\%-TPR that measures the false
positive rate (FPR) = FP/(FP+TN) when the true positive
rate (TPR) = TP/(TP+FN) is 95\%, where TP, TN, FP, and
FN denotes true positives, true negatives, false positives,
and false negatives, respectively. It can be interpreted as the probability that an example predicted incorrectly is misclassified as a correct prediction when TPR is equal to 95\%.

\begin{table}[!t]
    \begin{center}
		\resizebox{\columnwidth}{!}{
			\begin{tabular}{|l|c|c|c|c|c|c|c|}
				\hline \hline 
				\multirow{3}{*}{Backbones}& \multirow{3}{*}{Methods} &\multicolumn{2}{c|}{CIFAR10-LT  \cite{cui2019class}} &\multicolumn{2}{c|}{CIFAR100-LT \cite{cui2019class}}   \\ 
                & & \multicolumn{2}{c|}{IF=10} & \multicolumn{2}{c|}{IF=10} 
                \\
                    
                        &\multicolumn{1}{c|} {}&\multicolumn{1}{c|} {Acc. $\uparrow$}   &\multicolumn{1}{c|} {AURC $\downarrow$}
                        &\multicolumn{1}{c|}  {Acc. $\uparrow$} &\multicolumn{1}{c|} {AURC $\downarrow$}                                  
                        \\ \hline
			         
				    & MSP \cite{hendrycks2016baseline} & 88.49±0.18  & 40.96±3.19   & 59.39±0.23  &  196.28±3.57 \\
				      & RegMixup \cite{pinto2022using}& 91.28±0.15  & 17.74±0.99   & 62.51±1.13  &  156.56±4.06  \\
                        & CRL \cite{moon2020confidence}& 88.21±0.14  & 38.78±2.24  &60.33±0.29  & 181.33±3.63\\
\bf ResNet18 \cite{he2016deep} & SAM \cite{foret2020sharpness}& 88.56±0.38  &  27.44±1.39  & 60.24±0.44 & 183.68±3.17 \\
                        & SWA \cite{izmailov2018averaging}& 90.37±0.15  & 20.88±0.90   &63.86±0.11  & 157.43±1.63\\
                        & FMFP \cite{zhu2022rethinking}&90.46±0.06   & 18.55±0.35   &63.20±0.44  &153.88±1.91\\		
				    &  \bf SURE& \bf 92.65±0.11  & \bf 14.68±0.86   & \bf 66.83±0.38 & \bf 122.18±0.93\\
		              \hline \hline
				    & MSP \cite{hendrycks2016baseline} & 86.65±0.16  &84.26±4.55    &57.96±0.28  &257.81±1.84\\
				      & RegMixup \cite{pinto2022using}&89.53±0.30   & 26.75±0.39   &61.75±0.08  & 200.65±4.04 \\
                        & CRL \cite{moon2020confidence}& 86.45±0.21  & 87.05±1.79   & 57.69±0.25 &255.38±5.34\\
\bf VGG16-BN \cite{simonyan2014very} & SAM \cite{foret2020sharpness}& 88.24±0.51  &40.77±3.57    & 59.17±0.48 & 223.72±6.66\\
                        & SWA \cite{izmailov2018averaging}&89.23±0.05   & 25.02±0.66   & 60.95±0.51 &188.60±5.36\\
                        & FMFP \cite{zhu2022rethinking}& 89.23±0.22  & 21.55±0.34   &61.12±0.22  &179.68±1.90\\		
				    & \bf SURE& \bf 90.47±0.23  & \bf 19.51±0.59   & \bf 62.31±0.36 & \bf 158.17±2.43\\
		            \hline \hline
                        & MSP \cite{hendrycks2016baseline} & 87.75±0.53  &  37.94±7.71  &58.61±0.03  & 225.57±2.51\\
				      & RegMixup \cite{pinto2022using} &91.73±0.16   & 17.07±0.12   &65.14±0.10  & 131.85±1.81\\
                        & CRL \cite{moon2020confidence}&88.11±0.21   & 38.65±1.47   &60.06±0.15  &188.90±3.69 \\
\bf DenseNetBC \cite{huang2017densely}& SAM \cite{foret2020sharpness}& 88.79±0.29  & 27.02±1.23   &61.14±0.34  & 188.08±3.77\\
                        & SWA \cite{izmailov2018averaging}& 90.76±0.40  & 16.77±1.06   &64.52±0.75  &149.15±5.80\\
                        & FMFP \cite{zhu2022rethinking}&90.72±0.49   & 15.80±1.37   & \bf 65.62±0.24 &136.10±1.03 \\	
				    & \bf SURE& \bf 91.76±0.23  & \bf 13.72±0.72   & 65.34±0.08 & \bf 130.95±2.23\\
		              \hline \hline
                        & MSP \cite{hendrycks2016baseline}& 89.44±0.10  & 37.28±1.34   & 62.46±0.05 & 185.31±0.83\\
				      & RegMixup \cite{pinto2022using}&92.44±0.29   &  14.66±1.96  & 65.99±0.60  &144.91±3.02\\
                        & CRL \cite{moon2020confidence}&89.57±0.28   & 37.63±2.31   & 63.22±0.24 & 159.26±2.60\\
\bf WRNet28 \cite{zagoruyko2016wide}& SAM \cite{foret2020sharpness}& 90.86±0.13&21.11±0.72    & 65.27±0.13	 &145.33±2.15 \\
                        & SWA \cite{izmailov2018averaging}&92.17±0.27   &12.70±0.83    &68.73±0.17 &	122.27±1.09 \\
                        & FMFP \cite{zhu2022rethinking}&92.04±0.07   & 11.35±0.17   & 69.12±0.40 &111.44±1.31\\	
				    &\bf SURE& \bf93.91±0.01 & \bf 9.40±0.41  & \bf 70.92±0.27   & \bf 102.64±1.85 \\
		            \hline \hline
            
		    \end{tabular}}
		\caption{\textbf{Comparison of the performance of failure prediction on CIFAR10-LT and CIFAR100-LT \cite{cui2019class} with imbalance factor 10.} We keep 10\% training data as the validation set to select the best model. The means and standard deviations over \textit{three} runs are reported. $\downarrow$ and $\uparrow$ indicate that lower and higher values are better respectively. For each experiment, the best result is
shown in boldface. AURC \cite{geifman2018bias} values are multiplied by $10^{3}$ and all remaining values are in percentage. On datasets with long-tailed distributions, SURE outperforms other methods in almost all cases.}
		\label{tab::longtail_failure_prediction}
	\end{center}
\end{table}

\section{More results of failure prediction under distribution shift}
\label{section::corruption}

\begin{figure*}
\centering
\begin{subfigure}{\textwidth}
    \centering
    \includegraphics[width=\textwidth]{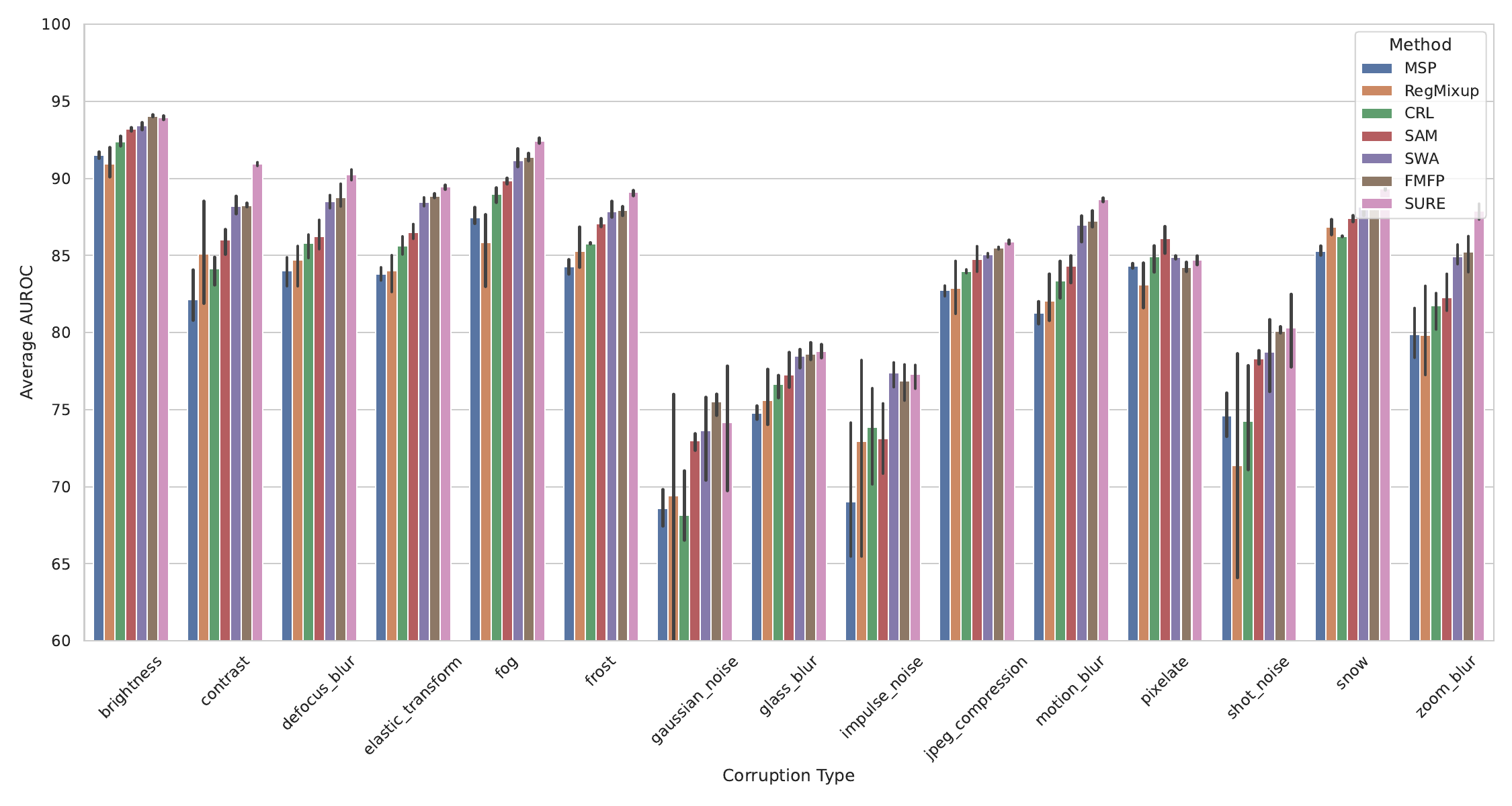}
    \caption{AUROC}
    \label{figure::cifarc acc}
\end{subfigure}
\begin{subfigure}{\textwidth}
    \centering
    \includegraphics[width=\textwidth]{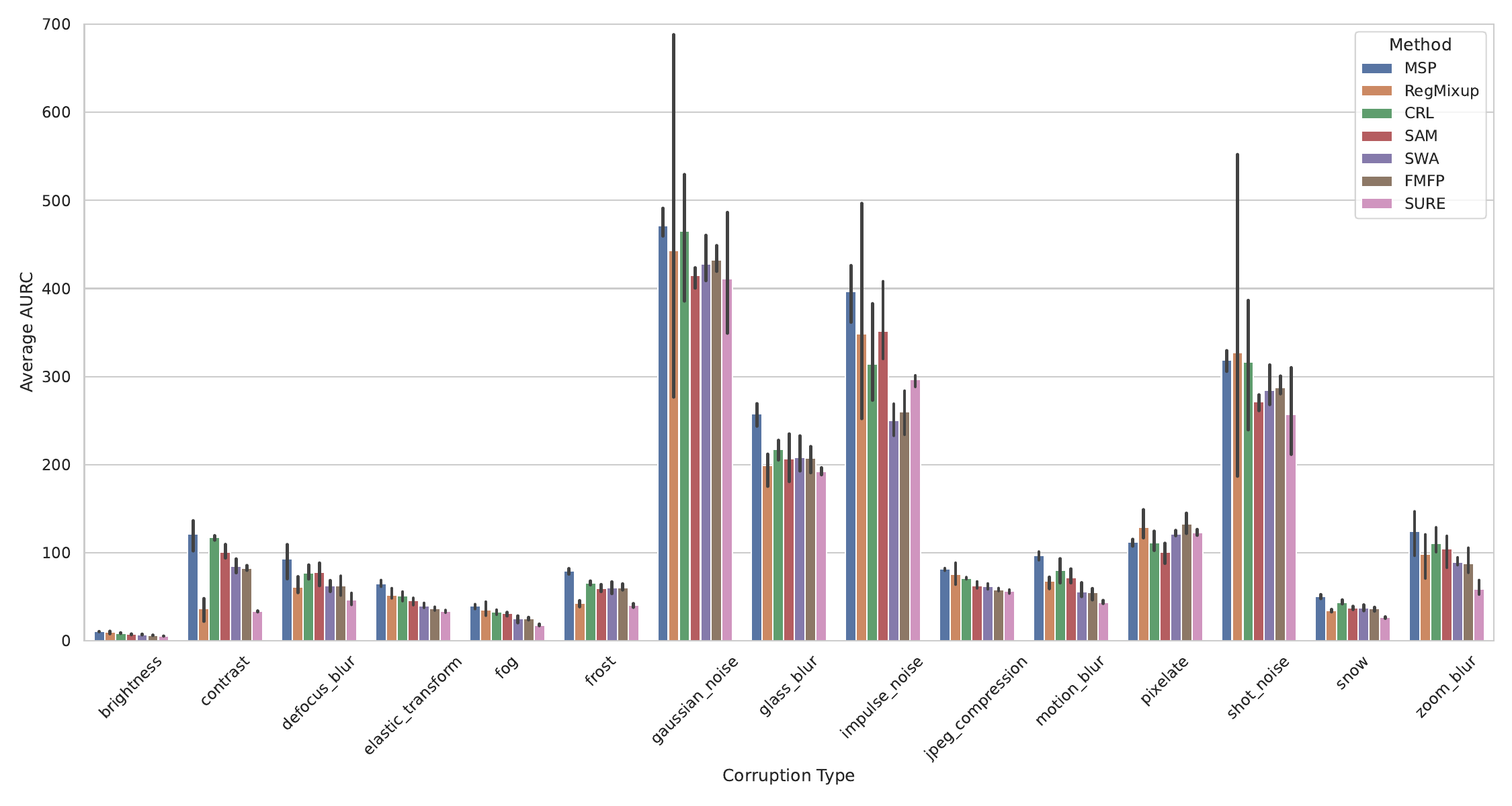}
    \caption{AURC}
    \label{fig::cifarc aurc}
\end{subfigure}
\caption{\textbf{Comparison of the average AUROC \cite{davis2006relationship} (higher is better) and AURC \cite{davis2006relationship} (lower is better) on CIFAR10-C \cite{hendrycks2019benchmarking}.} We choose DenseNet \cite{huang2017densely} as the backbone and CIFAR-10 as the training set. The evaluation results are averaged across the images with 5 severity levels under 15 types of corruption.}
\label{fig::each corruption}
\end{figure*}

In this section, we present the detailed performances of each corruption in Figure ~\ref{fig::each corruption}. We can observe that \textbf{SURE} outperforms the other methods in almost all corruption types. This consistent superiority across various corruption types indicates the robustness of \textbf{SURE}.
\section{Impact of different re-weighting maps}
\label{section::re-weighting}
In this section, we investigate the impact of different re-weighting maps on our uncertainty-aware re-weighting strategy in Table \ref{tab:re-weighting map}. Specifically, we explore four methods: exponential (exp), threshold, power, and linear.
Let \( s_i \) represent the confidence scores. We define these re-weighting methods with tuning parameters \( t \), \( \alpha \), and \( p \) as follows:

\begin{itemize}
    \item \textbf{Exponential: } The weights are defined using the exponential function:
    \[ \text{weights} = e^{-t \times s_i} \]
    where \( t \) is a scaling factor affecting the influence of confidence scores.

    \item \textbf{Threshold :}
    \[ \text{weights} = 
    \begin{cases} 
    1.0 - s_i , & \text{if }  s_i  < \alpha \\
    0, & \text{otherwise} 
    \end{cases} \]
    Here, \( \alpha \) is the threshold value.

    \item \textbf{Power: } The weights are determined by raising the term to a power:
    \[ \text{weights} = (1.0 - s_i )^p \]
    In this case, \( p \) is the exponent applied to the term \( 1.0 - s_i \) .

    \item \textbf{Linear: } A linear relationship is used to calculate the weights:
    \[ \text{weights} = 1.0 - s_i \]
    This method directly subtracts the confidence scores from 1.0.
\end{itemize}
 Based on the best result in Table \ref{tab:re-weighting map}, we choose ``exp'' (exponential) with t = 1 as the re-weighting map for all our long-tail classification experiments.
 
\begin{table}[!t]
    \centering 
    \resizebox{0.4\columnwidth}{!}{
        \begin{tabular}{|l|c|}
            \hline \hline 
            Methods & Acc. \\ 
            \hline
            w/o. re-weighting &  87.72 \\
            \hline \hline
            \multicolumn{2}{|c|}{exp} \\
            \hline
            $t = 0.5$ & 89.73 \\
            $t = 1$ & \bf{90.22} \\
            $t = 2$ & 88.96 \\
            \hline \hline
            \multicolumn{2}{|c|}{threshold} \\
            \hline
            $\alpha = 0.5$ & 89.35 \\
            $\alpha = 0.6$ & 89.50 \\
            $\alpha = 0.7$ & 89.01 \\
            $\alpha = 0.8$ & 89.60 \\
            $\alpha = 0.9$ & 89.87 \\
            \hline
            \hline
            \multicolumn{2}{|c|}{power} \\
            \hline
            $p = 2$ & 89.82 \\
            $p = 3$ & 89.44 \\
            $p = 4$ & 89.60 \\
            $p = 5$ & 89.25 \\
            \hline \hline
            linear & 89.60 \\
            \hline \hline
        \end{tabular}
    }
    \caption{\textbf{Impact of different re-weighting maps.} We have investigated the impact of different re-weighting maps on our uncertainty-aware re-weighting strategy on CIFAR10-LT \cite{cui2019class} with an Imbalance Factor (IF) of 50. Based on our findings, `exp' (exponential) method with $t = 1$ was selected as the re-weighting map for all our long-tailed classification experiments.}
    \label{tab:re-weighting map}
\end{table}

\end{document}